\pdfoutput=1
\documentclass[11pt]{article}
\usepackage{dsfont}
\newcommand{\Iverson}[1]{\mathds{1}({{#1}})}%

\usepackage[preprint]{acl}
\usepackage{multirow}
\usepackage{booktabs}
\usepackage{times}
\usepackage{latexsym}
\usepackage{amsmath}
\usepackage{lipsum}
\usepackage{tabularx}
\usepackage{array}
\usepackage[T1]{fontenc}
\usepackage[utf8]{inputenc}

\usepackage{microtype}

\usepackage{inconsolata}

\usepackage{graphicx}
\usepackage{colortbl} 
\usepackage{enumitem}

\title{Efficient Test-Time Scaling via Self-Calibration}

\author{
Chengsong Huang$^{1}$, Langlin Huang$^{1}$ \textbf{Jixuan Leng$^{2}$}\\
 \textbf{Jiacheng Liu$^{3}$}, \textbf{Jiaxin Huang$^{1}$}  \\
$^1$Washington Univeristy in St. Louis \\
$^2$Carnegie Mellon University
$^3$University of Washington \\
\texttt{\{chengsong,h.langlin,jiaxinh\}@wustl.edu} \\
\texttt{jixuanl@cs.cmu.edu, liujc@cs.washington.edu}}

\begin{document}
\maketitle
\begin{abstract}
Increasing test-time computation is a straightforward approach to enhancing the quality of responses in Large Language Models (LLMs). While Best-of-N sampling and Self-Consistency with majority voting are simple and effective,
they require a fixed number of sampling responses for each query, regardless of its complexity. This could result in wasted computation for simpler questions and insufficient exploration for more challenging ones.
In this work, we argue that model confidence of responses can be used for improving the efficiency of test-time scaling.
Unfortunately, LLMs are known to be overconfident and provide unreliable confidence estimation.
To address this limitation, we introduce \textbf{Self-Calibration} by distilling Self-Consistency-derived confidence into the model itself. This enables reliable confidence estimation at test time with one forward pass. 
% Our self-calibration framework generates input-output pairs from seed datasets and assigns confidence scores using soft self-consistency. It then trains the LLM with a combined loss to improve calibration while preserving its generation capability.
% \ljc{Here I think you should add a sentence to summarize how this self-calibration thing is done.}
We then design \textbf{confidence-based efficient test-time scaling methods} to handle queries of various difficulty, such as Early-Stopping for Best-of-N and Self-Consistency with calibrated confidence.
Experiments on three LLMs across six datasets demonstrate the effectiveness of our approach. 
% in both resource utilization and performance. 
Specifically, applying confidence-based Early Stopping to Best-of-N improves MathQA accuracy from 81.0 to 83.6 with a sample budget of 16 responses, 
indicating the efficency of the confidence-based sampling strategy at inference time~\footnote{Our codes are available at~\url{https://github.com/Chengsong-Huang/Self-Calibration}.}.
% across different questions, unlike Best-of-N, which uses a fixed number.

% \ljc{Mention some numbers from experiment results here.}
%result
\end{abstract}

\section{Introduction}

% \ljc{I have a nuanced high-level comment. Overall, the story I'm getting is that you have a better method to obtain confidence scores, and using this confidence score you can do efficient test-time scaling. I feel that these two things are less coupled, in a sense that efficient test-time scaling should be (qualitatively) doable with some sort of confidence score, and your better confidence score (quantitatively) makes the test-time scaling more efficient. I think an important baseline is how well you can do test-time scaling with the basic confidence score (i.e. Kadavath's P(True) itself). It would be nice to show that your improved confidence estimate is useful in application, since this improvement is motivated by the application in test-time scaling.}

%Increasing test-time computation is a straightforward approach to enhancing the quality of responses in LLMs. 
Leveraging additional computation during inference can enhance the quality of responses generated by large language models (LLMs)~\citep{Snell2024ScalingLT, Yao2023TreeOT, Wu2024InferenceSL, Chen2025SETSLS}. Among these methods, repeated sampling~\citep{Brown2024LargeLM} such as Best-of-N~\citep{Cobbe2021TrainingVT} and Self-Consistency~\citep{Wang2022SelfConsistencyIC} generate multiple candidate responses and select the final answer by a scoring model or a majority voting rule. While these methods have proven effective,
% However, their performance is highly sensitive to the number of samples generated, making hyperparameter selection crucial~\citep{Sessa2024BONDAL}. 
they require a fixed amount of sampled responses for each query regardless of its difficulty and complexity.
Although increasing the sample size generally improves performance, it also increases computational costs and inference time~\cite{Amini2024VariationalBA}. This is particularly inefficient for simple questions like “2 + 3 = ?”, where a few samples are  sufficient to find the correct solution~\cite{Chen2024DoNT}, and extensive sampling is unnecessary.
% and a fixed number of samples leads to unnecessary resource consumption.

\begin{figure}
    \centering
    \includegraphics[width=0.85\linewidth]{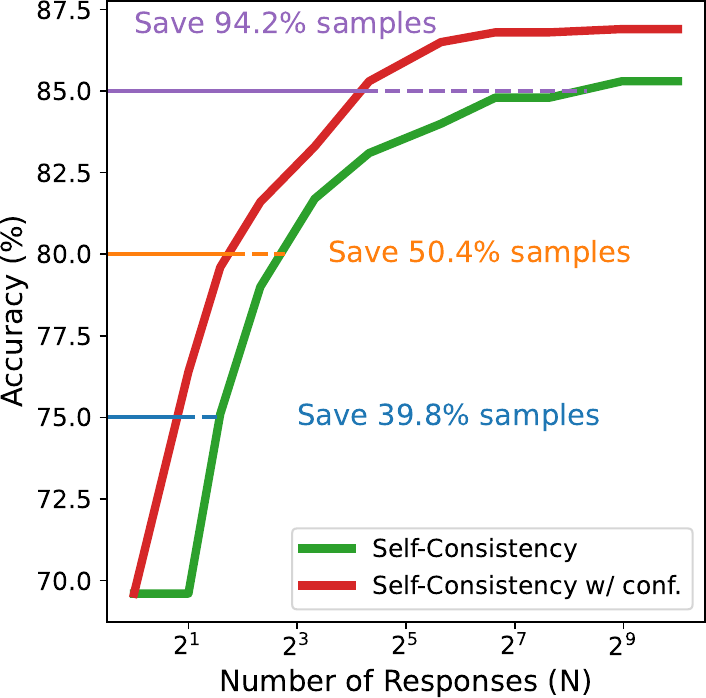}
    \caption{Accuracy over response numbers of standard Self-Consistency (SC) vs. confidence-weighted Self-Consistency (SC w/ conf.) on MathQA using our trained Llama-3.1-8B-Instruct model. The horizontal lines mark the response usage difference required for SC w/ conf. to reach the same accuracy with SC.}
    \vspace{-15pt}
    \label{fig:intro}
\end{figure}

% In this work, we focus on addressing this inefficiency in test-time scaling. 
% Previous methods have been proposed to adaptively determine the number of samples based on task complexity. 
Previous adaptive sampling strategies~\citep{Aggarwal2023LetsSS,Li2024EscapeSC,Wan2024ReasoningAS} typically design lightweight stopping criteria to determine whether additional responses should be sampled.
However, they often incorporate manually designed features or heuristic rules, such as stopping when the model generates the same response three times consecutively,
% ~\jh{can we add an example here? Otherwise, readers may be confused about how our early stopping differs from theirs}
which can limit their generalizability across different tasks and models. 
% Such handcrafted features typically require domain-specific tuning, reducing adaptability to diverse reasoning scenarios. 
Therefore, it is critical to design a task-independent, model-agnostic approach without heavy reliance on human-designed heuristics.

We propose an efficient test-time scaling method by using model confidence for dynamically sampling adjustment, since confidence can be seen as an intrinsic measure that directly reflects model uncertainty on different tasks. 
However, extracting accurate confidence can be challenging since LLMs are known to be overconfident on their own responses~\cite{lin2022teaching, xiong2023can, Leng2024TamingOI}, and their confidence often exceeds the actual accuracy. Self-Consistency~\cite{Wang2024SelfConsistencyBC} can provide a relatively accurate confidence estimation by aggregating answer counts from multiple sampled solutions~\cite{Tian2023JustAF}, but it again requires sampling a large number of responses for each query beforehand.
% However, existing calibration methods often require extra computational resources~\cite{Wang2024SelfConsistencyBC} or fail to provide sufficiently accurate confidence scores~\cite{Tian2023JustAF}, especially in scenarios where the performance of LLMs is suboptimal, requiring test-time scaling. 

To address this, we introduce \textbf{Self-Calibration} to train LLMs for accurate confidence estimation in only one forward pass, without requiring any human-labeled data. Specifically, we improve model calibration 
% leverage the model’s self-consistency score—a relatively accurate, but computationally expensive confidence estimation—as a training target. By doing so, we 
by distilling Self-Consistency-derived confidence into the model itself. This is done by constructing pseudo training tuples of query, answer, and confidence on a diverse training set.
At test time, we design \textbf{efficient test-time scaling strategies using these calibrated confidence scores}, such as early stopping for Best-of-N when sampled responses reach a target confidence, and Self-Consistency weighted by reliable confidence.
% To preserve its original generation ability, we also add the responses with high confidence as training data during the Self-Calibration, which is a strategy similar to Self-Improvement~\citep{Huang2022LargeLM} and SCPO~\citep{Prasad2024SelfConsistencyPO}.

%Building on efficiently produced confidence scores, we refine prior test-time scaling-up methods to better handle queries of various difficulty.
% Building on efficiently produced confidence scores, we refine prior test-time scaling-up methods to better handle queries of varying difficulty. Instead of relying on fixed heuristics or manually tuned thresholds, we leverage the model’s self-calibrated confidence to dynamically adjust the number of samples during inference. Specifically, when the confidence score is high, we reduce sampling to save computational resources, whereas for lower-confidence cases, we increase sampling to improve reliability. This score can also be used as the weight of each response when using voting-based methods like self-consistency.

%Experiments on two LLMs across several datasets demonstrate the efficiency of our approach in both resource utilization and performance.
Empirical experiments on three LLM architectures across six datasets demonstrate that our confidence-based test-time scaling approaches consistently outperform their baseline counterparts under the same sampling budget. Specifically, both Early Stopping for Best-of-N and confidence-weighted Self-Consistency improve MathQA accuracy over their baselines from 81.0 to 83.6 with an average sampling budget of 16 responses. More importantly, our approaches can achieve comparable performance with substantially fewer computational resources. As shown in Fig.~\ref{fig:intro}, confidence-weighted Self-Consistency can save 94.2\% samples to achieve an accuracy of 85.0, compared to standard Self-Consistency, demonstrating that reliable confidence estimation can significantly enhance the computational efficiency of test-time scaling.

% \begin{itemize}
%     \item We propose a self-calibration framework that enhances the model’s confidence estimation without requiring labeled data. By leveraging self-consistency as a training signal, our method improves model calibration with minimal additional computational cost.
    
%     \item We refine test-time scaling methods by dynamically adjusting the number of sampled responses using self-calibrated confidence scores. This enables adaptive sampling based on query difficulty and incorporates confidence-weighted voting to improve final predictions.

%     \item  Our comprehensive analysis reveals the applicability and efficiency of our methods. We show that our methods consistently outperform baselines under the same sample budget. We also demonstrate early stopping achieves optimal performance with an average sampling count of fewer than two.\jxc{isn't this redundant with last paragraph?}

% \end{itemize}

\section{Repeated Sampling}
\label{sec:basic}
Repeated sampling~\citep{Brown2024LargeLM} is a framework that generates multiple responses with Chain-of-Thought prompting~\citep{Wei2022ChainOT}, then uses a verifier to get the final results.
We will introduce three fundamental repeated sampling strategies, which aim to enhance response quality by selecting the most suitable answer from multiple generated candidates.
\subsection{Best-of-N}
For each input query $x$, multiple candidate responses $\{y_i\}$ are sampled, where \(1 \leq i \leq N\). A scoring function---such as an additional reward model or a confidence generator---assigns each response a score \(c_i = \text{Score}(y_i)\). The simplest selection strategy, known as Best-of-N~\cite{Cobbe2021TrainingVT}, chooses the response with the highest score as the final answer as  
$
\hat{y} = \arg\max\limits_{y}\;c_j
$.

\subsection{Self-Consistency}
Self-Consistency~\cite{Wang2022SelfConsistencyIC} selects the most frequent response among multiple sampled candidates. Given candidate responses \(\{y_1, y_2, \dots, y_N\}\), the final answer is determined by majority voting:
{
$$
\hat{y} \;=\;
\arg\max_{z}\;
\sum_{i=1}^{N}\;\Iverson{y_i = z}.
$$
}
This approach enhances robustness by aggregating diverse model outputs rather than relying on a single highest-scoring response.

\begin{figure*}[t]
    \centering
    \includegraphics[width=\linewidth]{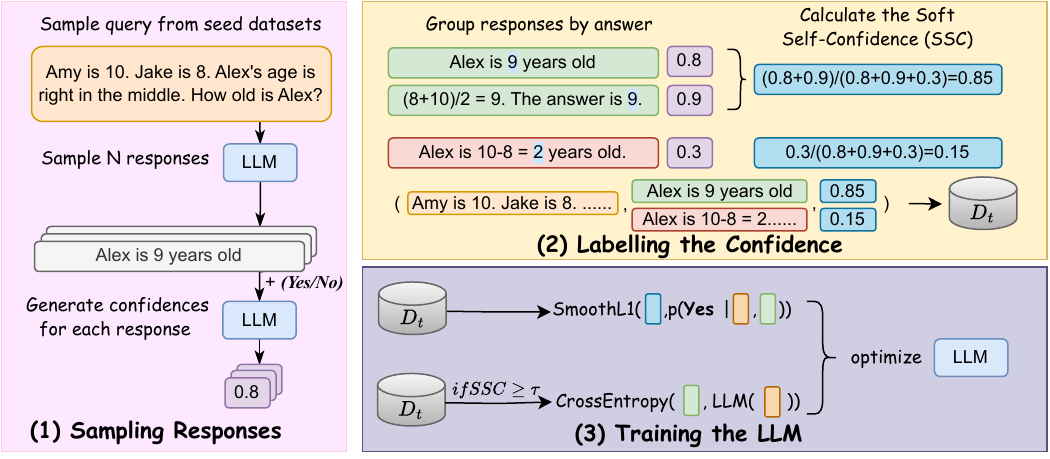}
    \caption{Illustration of the Self-Calibration framework. Given a query from the seed dataset, we sample $N$ responses from the LLM. We use a confidence querying prompt to let LLM assign a confidence score to each response. Responses are then grouped by their answers, and the Soft Self-Consistency (SSC) score is computed for each group. During training, all data tuples contribute to improving the model's calibration, while higher-confidence data is used to enhance the LLM's generation ability.}
    \vspace{-15pt}
    \label{fig:self-Calibration}
\end{figure*}

\subsection{Adaptive Self-Consistency}
Adaptive Self-Consistency (ASC)~\cite{Aggarwal2023LetsSS} enhances the standard Self-Consistency approach by dynamically adjusting the number of samples based on agreement among generated responses. This method iteratively samples responses and calculates the cumulative frequency \( v_k(z) \) and relative frequency \( \hat{r}_k(z) \) of each unique answer \( z \) after \( k \) samples:
% \vspace{-5pt}
\[
v_k(z) = \sum_{i=1}^k \Iverson{y_i = z}, \quad
\hat{r}_k(z) = \frac{v_k(z)}{k}.
\]

The sampling process continues until the maximum relative frequency \( \hat{r}_k(z) \) exceeds a predefined threshold \( \tau \). Formally:
\vspace{-5pt}
\[
\begin{cases}
    k \gets k+1, & \text{if } \max\limits_{z} \hat{r}_k(z) < \tau, \\
    y = \arg\max\limits_{z} \hat{r}_k(z), & \text{otherwise}.
\end{cases}
\]

This adaptive strategy reduces computational costs by limiting the number of required samples while maintaining high accuracy in the final answer selection.

\section{Self-Calibration}
\label{sec:self-cal}
In this section, we provide an overview of our proposed Self-Calibration framework, illustrated in
Fig.~\ref{fig:self-Calibration}. First, we synthesize a set of input-output-confidence tuples $(x_i, y_i, c_i)$ from a seed dataset for training, without requiring any ground-truth answer (Sec.~\ref{sec:data_gen}). Using this synthetic dataset, we can train a language model with a combined loss to output calibrated confidence scores (Sec.~\ref{sec:sctraining}). 
% Finally, we evaluate the calibration results to demonstrate the effectivessness of our proposed framework (Sec.~\ref{sec:sc_results}).

\subsection{Confidence Score Estimation}

% \ljc{This subsection makes it feel like you've already solved confidence scoring by this subsection. Perhaps start with ``A naive way to obtain confidence score is P(True) ...''}
A naive way to obtain a confidence score from LLM is $\operatorname{P}(\text{True})$~\citep{Kadavath2022LanguageM}.
Given the input-output pair $(x_i,y_i)$, we construct a prompt as $ x_i\oplus y_i\oplus I$, where $I$ is a confidence querying prompt, ``\text{Is the answer correct? (Yes/No)}''.
The confidence score is then defined as the probability of token ``\textbf{Yes}'' in the next position. $$c(x,y)=p_\theta (\textbf{Yes}|x,y,I)$$ Due to the KV-cache mechanism~\citep{Pope2022EfficientlyST}, the additional computational cost is roughly equivalent to generating 10 tokens, which is negligible compared to the typically longer input and output sequences. Empirical results suggest that $\operatorname{P}(\text{True})$ often lacks calibration, leading to overconfidence in incorrect answers~\cite{tian2023just}. So we aim to use supervised training to improve the calibration of $\operatorname{P}(\text{True})$, helping LLMs produce more reliable confidence scores.

\subsection{Training Data Generation}
% \jh{Here we want to emphasize that this data generation process does not need human annotated answer. Can you go over this subsection to refine it?}
\label{sec:data_gen}
% Here, we detail the process of generating training data $ (x,y,c)$. 
% The LLM first generates several different responses by dynamic temperature. Then we will use soft self-consistency to generate the confidence about $(x,y)$.~\jh{This paragraph is not a good summary because there is no logical chain. Instead of saying we do step 1, step 2, and step 3, we need to say, in order to fulfill goal 1, we do step 1; in order to fulfill goal 2, we do step 2. And the goals need to be related to your ultimate goal. This will make the paragraph more logical. Please go over all the method sections and make them more logical whenever you bring up a new technique/formula/equation.}

Our goal is to create a labeled dataset \(D_t = (x,y,c)_i\) without human annotations, 
where \((x,y)\) is a query--response pair and \(c\) is an accurate confidence. To achieve this, we first generate multiple candidate answers for each query and ensure diversity via Dynamic Temperature sampling. Next, we calibrate the confidence of each candidate through Soft Self-Consistency, which integrates the model's intrinsic probability estimate with the overall agreement among different responses.

\paragraph{Soft Self-Consistency Score.}
\label{sec:ssc}
Previous work has shown that self-consistency scores provide strong zero-shot calibration~\citep{Wang2024SelfConsistencyBC}, outperforming $\operatorname{P}(\text{True})$ or raw logits as confidence measures~\citep{Guo2017OnCO}. To further enhance the reliability of the confidence score in the training set, we introduce a soft self-consistency score, which integrates $\operatorname{P}(\text{True})$ with self-consistency and offers a more accurate and robust confidence estimation.

For each query $x$, we use the LLM to generate $N$ different responses, each with an associated confidence score. Given the set of triplets $ (x,y_n, c_n)$ where $1\leq n \leq N$, we compute the soft self-consistency (SSC) score as:
\[
\text{SSC}(y) = \frac{\sum_{i: y_i = y} c_i}{\sum_{i=1}^{N} c_i}.
\]
Using this score, we construct the final training set as $(x,y_i,\operatorname{SSC}(y_i))$, where $\operatorname{SSC}(y_i)$ provides a calibrated confidence estimation for each response.

\paragraph{Dynamic Temperature.}
\label{sec:dynam}
To generate more diverse and high-quality responses, we adopt the Entropy-based Dynamic Temperature (EDT) Sampling method~\citep{Zhang2024EDTIL} when generating each response \(y\).
By adaptively increasing the temperature when the entropy \(H\) of the output distribution is low, EDT promotes greater response diversity while preserving output quality.
Formally, the temperature \(T(H)\) is defined as:
\vspace{-5pt}
\[
T(H) =
\begin{cases}
T_0 \times M^{\gamma / H}, & \text{if } T_0 \times M^{\gamma / H} \ge \tau_0, \\
0, & \text{otherwise},
\end{cases}
\]
where \(T_0\) is the base temperature, \(M\) is a scaling factor, \(\gamma\) affects the scale of temperature variations, and \(\tau\) is a threshold that sets the temperature to zero if \(T_0 \times M^{\gamma / H}\) is below \(\tau_0\).

% ~\jh{explain these symbols, note that $\theta$ is already used as LLM parameter so we can replace the $\theta$ here with sth. else.}

\subsection{Training Objective}
\label{sec:sctraining}
We optimize the model's confidence estimation by minimizing the difference between the predicted confidence and the target confidence using the SmoothL1 loss. To ensure that training on confidence estimation does not degrade the model's reasoning ability, we incorporate the standard generation loss of Chain-of-Thought answers into the objective~\cite{Huang2022LargeLM}. Specifically, only responses with confidence scores above a threshold $\eta$ are selected for training to guarantee the quality of the reasoning path. A weighting coefficient $w$ is introduced to balance these two loss terms. The overall loss function is formulated as:
\[
\small
\begin{aligned}
\mathcal{L}_{\text{total}}(\theta)
&=\;\sum_{(x_j,\,y_j)\in \mathcal{D}}
\mathrm{SmoothL1}\Bigl(
\,p_\theta(\operatorname{Yes}\mid x_j,y_j,I),\; c_j
\Bigr)\,\\
&\quad +\omega \sum_{\substack{(x_i,\,y_i) \\ c_i > \eta}}
\Bigl(
-\log\,p_\theta\bigl(y_i \,\bigm\vert\, x_i\bigr)
\Bigr).
\\
\end{aligned}
\]

\section{Confidence-Guided Test-Time Scaling}
\label{sec:refined}
We then introduce how to incorporate reliable confidence scores obtained from Self-Calibration to existing test-time scaling methods.
% In this section, we discuss how the model's confidence score can enhance efficient repeated sampling methods. First, we present refined versions of the methods introduced in Sec.~\ref{sec:basic} by incorporating the confidence score. 
% Then, we will introduce a new metric, Relative Selection Gain, to evaluate the performance of each method (Sec.~\ref{sec:metric}).

% \subsection{Refined Version}

\subsection{Early Stopping with Confidence}
Early Stopping improves the efficiency of Best-of-N by dynamically terminating the sampling process once a response with sufficient confidence is found. Given a sequential sampling process where each response \( y_i \) is assigned a confidence score \( c_i \), we follow this rule:

\[
\begin{cases}
    k \gets k+1, & \text{if } c_i < \tau, \\
    y = y_i, & \text{otherwise}.
\end{cases}
\]

This means that we continue sampling responses one by one until a response meets the confidence threshold \( \tau \), and such a response is selected as the final answer, avoiding unnecessary additional sampling and reducing computational overhead.

\subsection{Self-Consistency with Confidence}  
Self-Consistency with Confidence extends the traditional Self-Consistency approach by incorporating confidence scores into the voting process. Instead of treating all sampled responses equally, we assign each response \( y_i \) a confidence score \( c_i \), leading to a weighted aggregation:
\vspace{-5pt}
\[
y \;=\;
\arg\max_{z}\;\sum_{i=1}^{N}\;c_i\,\Iverson{\,y_i = z\,}.
\]

This modification ensures that responses with higher confidence contribute more significantly to the final selection, enhancing robustness by prioritizing more reliable predictions.

\subsection{Adaptive Self-Consistency with Confidence}
Similar to Self-Consistency with Confidence, we use confidence as the weight when calculating the relative frequency in Adaptive Self-Consistency. 
% Then we will follow the rest part in Adaptive Self-Consistency.
\[
v_k(z) = \sum_{i=1}^k c_i\Iverson{y_i = z}, \quad
\hat{r}_k(z) = \frac{v_k(z)}{\sum_{i=1}^k c_i}.
\]

% \subsection{Relative Selection Gain}
% \label{sec:metric}
% To better evaluate the efficiency of different test-time scaling methods, we propose a new metric, Relative Selection Gain (\metric).

% We first revisit the upper bound of methods that rely on repeated sampling. Suppose we have an oracle selector capable of identifying the correct answer from candidate responses. The performance in this idealized setting is given by $\text{pass@}N$~\citep{Chen2021EvaluatingLL}:
% \[
% \text{pass@}N =
% \frac{1}{|\mathcal{D}|}
% \sum_{x \in \mathcal{D}}
% \Iverson{\exists i \in [1, N], y_i \text{ is correct}}.
% \]
% To quantify the effectiveness of test-time scaling, we measure its relative performance compared to $\text{pass@}N$. Specifically, we define \metric\space as:
% \[
% \text{RSG}_N = \frac{\text{Acc@}N - \text{pass@}1}{\text{pass@}N - \text{pass@}1},
% \]

% This metric captures how effectively a method recovers the maximum possible accuracy gain. A higher \metric\space indicates better utilization of multiple samples, approaching oracle-level performance.

\section{Experiments}
\begin{table*}[th]
\centering
\small
\begin{tabular}{ll|cc|cc|cc}
\toprule
\multirow{2}{*}{\textbf{Dataset}} & \multirow{2}{*}{\textbf{Metric}} & \multicolumn{2}{c|}{\textbf{Llama-3.1-8B-Instruct}} & \multicolumn{2}{c|}{\textbf{Qwen2.5-7B-Instruct}} & \multicolumn{2}{c}{\textbf{DeepSeek-R1-Distill-1.5B}} \\\cmidrule(lr){3-4} \cmidrule(lr){5-6} \cmidrule(lr){7-8}
 & & \textbf{Vanilla} & \textbf{Self-Calibration} & \textbf{Vanilla} & \textbf{Self-Calibration} & \textbf{Vanilla} & \textbf{Self-Calibration} \\
\midrule
\multicolumn{8}{l}{\textit{In-Domain Datasets}} \\
 & ECE $\downarrow$ & 13.70 & \textbf{3.79} & 87.39 & \textbf{16.88} & 46.66 & \textbf{40.03} \\
 GSM8K & AUC $\uparrow$ & 72.43 & \textbf{75.36} & 68.61 & \textbf{82.21} & \textbf{64.31} & 55.57 \\
 & ACC $\uparrow$ & 77.44 & \textbf{80.43} & \textbf{89.41} & 88.74 & 73.38 & \textbf{75.36} \\
\midrule
 & ECE $\downarrow$ & 28.03 & \textbf{10.17} & 89.60 & \textbf{24.49} & 30.40 & \textbf{12.00} \\
 SVAMP & AUC $\uparrow$ & 74.17 & \textbf{75.79} & 75.10 & \textbf{87.46} & 49.33 & \textbf{71.27} \\
 & ACC $\uparrow$ & 72.60 & \textbf{75.29} & 90.48 & \textbf{92.00} & 52.27 & \textbf{57.48} \\
\midrule
 & ECE $\downarrow$ & 5.45 & \textbf{5.00} & 57.58 & \textbf{5.62} & 20.19 & \textbf{11.36} \\
 ARC\_easy & AUC $\uparrow$ & \textbf{81.16} & 76.89 & 66.10 & \textbf{76.75} & 62.89 & \textbf{66.86} \\
 & ACC $\uparrow$ & 87.73 & \textbf{89.21} & \textbf{92.11} & 92.01 & 54.00 & \textbf{56.74} \\
\midrule
\midrule
\multicolumn{8}{l}{\textit{Out-of-Domain Datasets}} \\
 & ECE $\downarrow$ & 7.01 & \textbf{6.03} & 55.19 & \textbf{10.11} & 11.42 & \textbf{5.46} \\
 ARC\_challenge & AUC $\uparrow$ & \textbf{80.67} & 80.41 & 64.21 & \textbf{78.33} & 64.07 & \textbf{65.27} \\
 & ACC $\uparrow$ & 80.87 & \textbf{82.38} & \textbf{89.37} & 89.05 & 43.39 & \textbf{45.77} \\
\midrule
 & ECE $\downarrow$ & 27.85 & \textbf{9.69} & 72.41 & \textbf{5.82} & 47.26 & \textbf{4.60} \\
 Object Counting & AUC $\uparrow$ & 53.84 & \textbf{59.47} & 68.07 & \textbf{81.02} & 50.39 & \textbf{67.61} \\
 & ACC $\uparrow$ & 60.68 & \textbf{65.88} & 72.41 & \textbf{74.22} & 55.33 & \textbf{58.13} \\
\midrule
 & ECE $\downarrow$ & 12.55 & \textbf{8.64} & 62.35 & \textbf{18.92} & 13.16 & \textbf{4.34} \\
 MathQA & AUC $\uparrow$ & 85.23 & \textbf{87.21} & \textbf{72.48} & 69.80 & \textbf{78.89} & 66.09 \\
 & ACC $\uparrow$ & 44.18 & \textbf{52.34} & 49.85 & \textbf{64.18} & 37.69 & \textbf{43.21} \\
\bottomrule
\end{tabular}
\caption{Self-Calibration results across both in-domain and out-of domain datasets on three different models.}
\vspace{-15pt}
\label{tab:sc_results}
\end{table*}
\subsection{Experiment Setup}
\paragraph{Models.}
To evaluate our self-calibration framework and our efficient test-time scaling methods, we conduct experiments on three open-source LLMs: Llama-8B-3.1-Instruct~\footnote{\url{https://huggingface.co/meta-llama/Llama-3.1-8B-Instruct}}~\citep{Dubey2024TheL3}, Qwen2.5-7B-Instruct~\footnote{\url{https://huggingface.co/Qwen/Qwen2.5-7B-Instruct}}~\citep{qwen2.5} and DeepSeek-R1-Distill-Qwen-1.5B~\footnote{\url{https://huggingface.co/deepseek-ai/DeepSeek-R1-Distill-Qwen-1.5B}}~\citep{deepseekai2025deepseekr1incentivizingreasoningcapability}.
These models represent diverse architectures and training strategies, allowing us to test the adaptability of our methods. 
% By using open-source models in evaluation, we ensure the reproducibility of our proposed methods and validate their broad applicability across different model architectures.
% \jxc{I do not see the point of this last sentence, it feels redundant as well, since applicability is already mentioned in the previous sentences.}

\paragraph{Seed Datasets.}
We construct our training dataset with diverse reasoning datasets, including: ARC\_easy~\citep{allenai:arc}, commonsense QA~\citep{talmor-etal-2019-commonsenseqa}, LogiQA~\citep{Liu2020LogiQAAC}, GSM8K~\citep{cobbe2021gsm8k}, OpenBookQA~\citep{Mihaylov2018CanAS}, ReClor~\citep{Yu2020ReClorAR}, SciQ~\citep{Welbl2017CrowdsourcingMC}, SVAMP~\citep{patel-etal-2021-nlp} and WindGrande~\citep{Sakaguchi2019WinoGrande}. For each dataset, we randomly sample 2,000 questions from the training set to construct our training data. Additional details are shown in Appendix~\ref{app:hyperparameter}.

\paragraph{Evaluation Datasets and Prompts.}
We evaluate our methods on three benchmark datasets: ARC-Challenge~\citep{allenai:arc}, Object-Counting~\citep{Suzgun2022ChallengingBT} and MathQA~\citep{Amini2019MathQATI}, covering mathematical and commonsense reasoning tasks in both multiple-choice and open-ended formats. ARC-Challenge includes difficult science questions requiring external knowledge and reasoning. Object-Counting focuses on numerical and spatial reasoning by counting objects in various contexts. MathQA tests mathematical problem-solving across arithmetic, algebra, and calculus. 

These three datasets are considered out-of-domain as they differ from the datasets used in training, which we refer as in-domain datasets. To evaluate performance in an in-domain setting, we also use the test sets of GSM8K, SVAMP, and ARC\_easy.
% These datasets provide a diverse and rigorous evaluation of our approach.
The system prompt and the task prompt of each dataset are shown in Appendix~\ref{app:prompt}.
% \paragraph{Hyperparameter}
% When evaluating the performance of t
\paragraph{Baseline Methods.}
In addition to the repeated sampling methods mentioned in Sec.~\ref{sec:basic}, we also include other adaptive test-time scaling methods such as Early-Stopping Self-Consistency (ESC)~\cite{Li2024EscapeSC} and Reasoning-Aware Self-Consistency (RASC)~\cite{Wan2024ReasoningAS} for comparison. 
% Instead of generating a predetermined number of reasoning paths, 
ESC divides the sampling process into sequential windows and halts further sampling when a high-confidence consensus is reached within a window.
RASC enhances sampling efficiency by dynamically evaluating both the generated answers and their corresponding reasoning paths. 
% Since ESC and RASC also aim to improve the efficiency of self-consistency, we include them as baseline methods to analyze the effectiveness of our approach in enhancing reasoning efficiency and accuracy.

% By integrating ESC and RASC as baseline methods, we aim to evaluate their effectiveness in improving reasoning efficiency and accuracy in our specific context.~\jh{Instead of saying we want to evaluate these baseline methods to test their effectiveness, we need to claim why we want to compare our methods against these baselines.}
\begin{table*}[h]
 \centering
 \small
 \setlength{\tabcolsep}{1.5pt}
 \resizebox{\textwidth}{!}{%
 \begin{tabular}{llll|lll|lll}
 \toprule
 & \multicolumn{3}{c}{Llama-3.1-8B-Instruct} & \multicolumn{3}{c}{Qwen2.5-7B-Instruct} & \multicolumn{3}{c}{DeepSeek-R1-Distill-1.5B} \\
 \cmidrule(lr){2-4} \cmidrule(lr){5-7} \cmidrule(lr){8-10}
 Methods & Obj\_C. & MathQA & ARC\_C. & Obj\_C. & MathQA & ARC\_C. & Obj\_C. & MathQA & ARC\_C. \\
 \midrule
 \rowcolor{gray!20} Pass@1 & 67.6 & 71.5 & 82.8 & 76.8 & 82.9 & 88.5 & 61.2 & 89.9 & 58.2 \\ 
 \midrule
 SC & 76.0 & 81.0 & 87.1 & 81.2 & 86.3 & \textbf{91.2} & \textbf{70.8} & 91.6 & 65.6 \\
 SC w/ Conf.* & \textbf{76.8} (+0.8) & 83.4 (+2.4)  & 87.4 (+0.3) & 80.8 (-0.4) & 87.5 (+1.2) & 90.5 (-0.7) & \textbf{70.8} (0.0) & \textbf{91.8} (+0.2) & 65.9 (+0.3) \\
 \rowcolor{blue!10}SC w/ Conf.  & \textbf{76.8} (+0.8) & \textbf{83.6} (+2.6) & \textbf{87.7} (+0.6) & 81.2 (0.0) & \textbf{87.8} (+1.5) & 90.8 (-0.4) & \textbf{70.8} (0.0) & \textbf{91.8} (+0.2) & \textbf{66.5} (+0.9) \\
 Best-of-N & 69.2 & 81.0 & 86.4 & 76.8 & 86.8 & 90.2 & 54.0 & 90.0 & 58.9 \\
 Early Stopping* & \textbf{76.8} (+7.6) & 83.4 (+2.4) & 87.3 (+0.9) & 80.8 (+4.0) & 87.5 (+0.7) & 90.5 (+0.3) & 64.8 (+10.8) & 91.6 (+1.6) & 65.9 (+7.0)\\
 \rowcolor{blue!10}Early Stopping & \textbf{76.8} (+7.6) & \textbf{83.6} (+2.6) & \textbf{87.7} (+1.3) & 81.2 (+4.4) & \textbf{87.8} (+1.0) & 90.8 (+0.6) & \textbf{70.8} (+16.8) & 91.6 (+1.6) & \textbf{66.5} (+7.6) \\
 ASC & 74.8 & 80.0 & 86.5 & \textbf{81.6} & 86.2 & 90.6 & 70.4 & 91.6 & 64.3 \\
 ASC w/ Conf.* & 74.8 (0.0) & 81.6 (+1.6) & 86.6 (+0.1) & \textbf{81.6} (0.0) & 86.9 (+0.7) & 90.4 (-0.2) & 70.4 (0.0) & 91.6 (0.0) & 64.7 (+0.4) \\
 \rowcolor{blue!10}ASC w/ Conf. & 75.2 (+0.4) & 81.9 (+1.9) & 86.6 (+0.1) & \textbf{81.6} (0.0) & 87.2 (+1.0) & \textbf{91.2} (+0.6) & 70.4 (0.0) & \textbf{91.8} (+0.2) & 65.1 (+0.8) \\
 ESC & 76.0 & 81.0 & 87.1 & 81.2 & 86.3 & {91.0} & \textbf{70.8} & 91.3 & 65.6 \\
 RASC & 76.0 & 81.4 & 87.3 & 81.2 & 86.4 & 90.3 & \textbf{70.8} & 91.4 & 65.8 \\
 \bottomrule
 \end{tabular}
 }
 \caption{
 Accuracy comparison of different test-time scaling methods across three language models when the sample budget equals to 16. The evaluation is conducted on three datasets: Obj\_C. (Object\_Counting), MathQA, and ARC\_C. (ARC\_Challenge). ``Sample budget'' refers to the average number of responses sampled per query. The improvements of confidence-augmented methods over their baselines are shown in parentheses. All methods use the same responses generated by Self-Calibration trained models, while methods marked with \textbf{*} use confidence scores from the vanilla model. The results when the sample budget equals 4 are shown in Appendix~\ref{app:full_table}. 
 }
 \vspace{-15pt}
 \label{tab:results}
\end{table*}
\subsection{Evaluation on Self-Calibration}
\label{sec:sc_results}
\paragraph{Evaluation Metrics.} 
We first evaluate how well our Self-Calibration approach enable models to output accurate confidence estimation.
We adopt three standard metrics for evaluating model calibration: Expected Calibration Error (ECE)~\citep{guo2017calibration}, Area Under the Receiver Operating Characteristic Curve (AUC)~\citep{hendrycks2016baseline}, and accuracy (ACC) on both in-domain and out-of-domain datasets. ECE measures the discrepancy between a model’s predicted confidence and its actual accuracy, defined as:
\vspace{-5pt}
\[
\text{ECE} = \sum_{m=1}^{M} \frac{|B_m|}{N} \left| \text{acc}(B_m) - \text{conf}(B_m) \right|,
\]
where $M$ is the number of bins, $B_m$ represents the set of samples in the $m$-th bin, and $N$ is the total number of samples.
A lower ECE value indicates better calibration, meaning the model’s confidence aligns more closely with its actual correctness.

\paragraph{Results.}

In Table~\ref{tab:sc_results}, 
% we present the results of our self-calibration framework across multiple in-domain and out-of-domain datasets. 
we compare our models trained on Self-Calibration objective with their vanilla base models on multiple in-domain and out-of-domain datasets. Self-Calibration trained models consistently lower the ECE score while generally improve accuracy. On GSM8K, Self-Calibration reduces ECE from 13.70 to 3.79 while improving accuracy from 77.44\% to 80.43\%. Even in cases where ECE slightly increases, such as ARC\_easy for Llama-3.1-8B-Instruct, accuracy still improves from 87.73\% to 89.21\%. Moreover, the strong results on out-of-domain tasks demonstrate the generalizability of our method, as seen in MathQA, where accuracy improves from 49.85\% to 64.18\% for Qwen2.5-7B-Instruct.

\paragraph{Ablation Study.}
\begin{table}[t]
\centering
\small
%\begin{tabular}{lcc|cc}
\begin{tabularx}{\columnwidth}{l c c | c c}
\toprule
 & \multicolumn{2}{c}{\textbf{MathQA}} & \multicolumn{2}{c}{\textbf{Object Counting}} \\
 \cmidrule(lr){2-3} \cmidrule(lr){4-5}
\textbf{Method} & \textbf{ECE $\downarrow$} & \textbf{ACC $\uparrow$} & \textbf{ECE $\downarrow$} & \textbf{ACC $\uparrow$} \\
\midrule
ours (full) & 8.64 & 52.34 & 9.69 & 65.88 \\
\hspace{0.5em}w/o EDT & 9.14 & 51.44 & 10.40 & 62.88 \\
\hspace{0.5em}w/o SSC & 10.85 & 52.18 & 16.02 & 61.12 \\
\hspace{0.5em}w/o L1-smooth & 6.43 & 50.86 & 10.48 & 56.48 \\
\bottomrule
%\end{tabular}
\end{tabularx}
\caption{Ablation study results on MathQA and Object Counting in Llama-3.1-8B-Instruct. ``w/o L1-smooth'' means using MSE loss instead of L1-smooth.}
\vspace{-15pt}
\label{tab:ablation}
\end{table}

% To further analyze the impact of each component in our Self-Calibration framework, 
We conduct an ablation study
% study on the MathQA and Object Counting datasets. Specifically, we 
to investigate the impact of key components in Self-Calibration, including Dynamic Temperature (EDT), Soft Self-Consistency (SSC), and L1-smooth loss. 
Table~\ref{tab:ablation} presents our ablation results on the MathQA and Object Counting datasets. 
% Our tuned Llama-3.1-8B-Instruct model achieves the best balance of low ECE and high accuracy, underscoring the effectiveness of both the dynamic temperature and soft self-consistency score components. 
Removing the dynamic temperature or the soft self-consistency score leads to noticeable increases in ECE and/or drops in accuracy. 
Meanwhile, replacing the L1-smooth objective with MSE achieves slightly lower ECE on MathQA but reduces accuracy on both tasks, suggesting that our chosen loss formulation is more robust overall.
These results demonstrate that each module contributes to model calibration and reasoning performance.

\subsection{Evaluation on Efficient Test-time Scaling}
% \jh{I think in sec. 5.2, we are comparing an uncalibrated model and a calibrated model, but in sec. 5.3, we are comparing different scaling strategies on the same calibrated models? That means we need to clearly convey this signal to the readers, to convince them that this is fair comparison. Please clearly explain this.} \ljc{To add to Jiaxin's point, I want to clarify if the basic scaling method (e.g. SC) uses the untuned model, while your confidence-weighted scaling method (e.g. SC w/ conf) uses the calibration-tuned model? If the underlying models are different, it might give your scaling method an unintended advantage, because the calibration-tuned model is trained on more targeted datasets (I understand you have in-domain and out-of-domain, but ARC-easy is pretty targeting for ARC-challenge). To ensure fair comparison, you might need to tune the default model with a supervised method on the same in-domain datasets.}

To ensure fair comparison across different test-time scaling methods, we use the same sample budgets for each of them. Sample budget refers to the average number of responses each method samples per query. 
% A higher sample budget typically enables more thorough exploration, but it also increases computational cost. 
For dynamic methods such as Early Stopping and ASC w/ Conf., we set internal thresholds so that the actual number of samples collected in practice is close to a target budget. To ensure a fair comparison, all methods use responses sampled from Self-Calibration trained models.

Table~\ref{tab:results} shows the accuracy comparison of different methods with a sample budget of 16. We observe that SC w/ Conf., Early Stopping, and ASC w/ Conf. consistently outperform their base counterparts. 
% For example, with a budget of 4 on Llama-3.1-8B-Instruct, SC w/ Conf. improves over SC on MathQA (78.8 to 82.1), while ASC w/ Conf. outperforms ASC (80.0 to 81.9). Similar trends appear with a budget of 16. 
On Llama-3.1-8B-Instruct, SC w/ Conf. surpasses SC on MathQA (81.0 to 83.6), while on DeepSeek-R1-Distill-1.B,
Early Stopping outperforms Best-of-N on ARC\_challenge (58.9 to 66.5). These results highlight that integrating calibrated confidence enhances test-time scaling with the same sampling budget. 
We also compare our approach with methods that use uncalibrated confidenc scores from the vanilla model (indicated by *). These methods generally underperform confidence from Self-Calibration trained model, indicating the necessity of confidence calibration. The results when the sample budget equals 4 are shown in Appendix~\ref{app:full_table}.

% At a budget of 4, each dynamic method yields higher accuracy by quickly filtering out or stopping unpromising samples, while at a budget of 16, allowing for more thorough sampling further improves final performance. 

% ~\jh{thorough sampling improves performance -- this is why we need test-time scaling, but how is this related to the paper's goal and claim? I am confused by the discussion of budget of 4 and 16 here.}
% These results confirm that incorporating confidence estimation and dynamic stopping mechanisms provides consistent gains over static approaches, and that allowing more sampling generally leads to stronger overall accuracy.~\jh{Make it clear here as well.}

\section{Analysis}

\subsection{Confidence Score Compared to Reward Score from Reward Models}
We compare our self-generated confidence scores with established open-source reward model approaches.
A reward model is an additional scoring model used to evaluate the quality of generated responses~\citep{Christiano2017DeepRL}. Deployment of a reward model can introduce several limitations: (1) Reward scores are often unbounded or require dataset-specific normalization, thus difficult to apply a universal threshold for filtering or reweighting responses; (2) Running an extra reward model increases inference time; and (3) A dedicated reward model requires additional GPU memory, and is less efficient for large-scale deployment.

For our analysis, we use the following reward models for comparison:
for Llama-3.1-Instruct,
we use the reward model from RLHFlow~\footnote{\url{https://huggingface.co/RLHFlow/Llama3.1-8B-ORM-Mistral-Data}}~\citep{dong2024rlhf}; for Qwen-2.5, we utilize its official Process Reward Model (PRM)~\footnote{\url{https://huggingface.co/Qwen/Qwen2.5-Math-PRM-7B}}~\citep{prmlessons}. 
For PRM, we use the lowest reward score across all steps. 
We ensure the size of each reward model matches with their corresponding base models. 

% ~\jh{Can we explicitly mention the reward model size here, instead of only in the url? We can also claim that our method is comparable to reward model of what size.}

\begin{table}[]
\centering
\small
\begin{tabularx}{\columnwidth}{Xlrr}
\toprule
Model & Dataset  & Reward & Confidence \\ 
\midrule
      & MathQA   & 82.1   & \textbf{84.0}  \\ 
Llama & Object Counting  & \textbf{72.6}  & 72.0   \\ 
      & ARC\_Challenge  & 86.2   & \textbf{86.6}  \\ 
      & MathQA   & \textbf{87.5}  & 86.8   \\ 
Qwen  & Object Counting  & \textbf{76.6}  & 76.4   \\ 
      & ARC\_Challenge  & 89.6   & \textbf{89.8}   \\ 
\bottomrule
\end{tabularx}
\caption{Accuracy of Best-of-16 on two models (Llama-3.1-8B-Instruct and Qwen-2.5-7B-Instruct) on three datasets between self-generated confidence scores and reward scores from additional reward models.}
\vspace{-15pt}
\label{tab:performance_comparison_re}
\end{table}

Table~\ref{tab:performance_comparison_re} shows that our self-generated confidence scores achieve similar performance to reward model scores across all datasets when using Best-of-N. This means that our method, by generating approximately 10 additional tokens, achieves a performance comparable to that of an extra reward model of the same size.

\subsection{Performance Comparison Under Different Sample Budgets}

\begin{figure}
    \centering
    \includegraphics[width=0.95\linewidth]{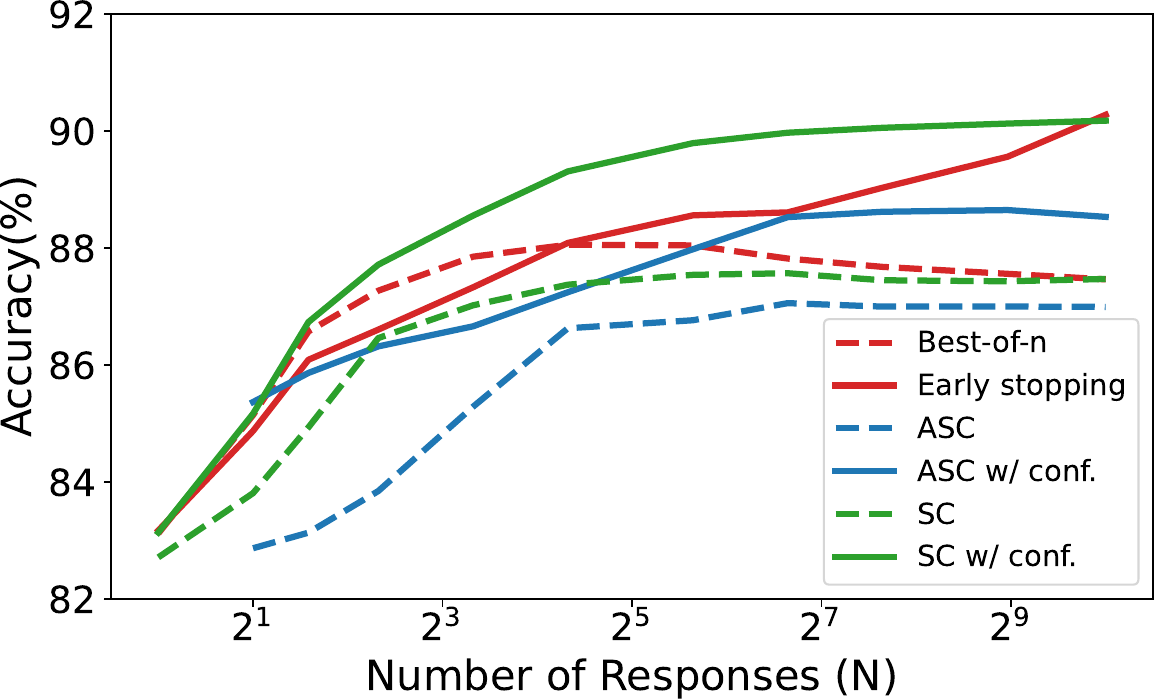}
    \caption{Accuracy over varying sample budgets of different inference strategies on MathQA using Self-Calibration trained Qwen-2.5-7B-Instruction. The results of other models and datasets are shown in Appendix~\ref{app:different_budgets}.}
    \vspace{-15pt}
    \label{fig:vary-budget}
\end{figure}

% In this section, we analyze how different methods perform under varying sample budgets. 
Increasing the sample budget allows for selecting higher-quality outputs but comes at the cost of greater computational expense. To evaluate this trade-off, we compare different methods across multiple sample budgets and visualize their performance trends.
As shown in Figure~\ref{fig:vary-budget}, all methods achieve better accuracy as the number of responses increases. Our confidence-guided approaches consistently outperform their original counterparts in most settings. When the sample budget is small, Best-of-N performs better than early stopping because early stopping might stop too soon with a low threshold, missing a better response.

% As shown in Figure~\ref{fig:enter-label}, all methods improve their final performance with larger sample budgets. Our confidence-guided method outperforms its original counterpart in most settings. Notably, by applying a threshold for early stopping, it achieves an average sample budget between 1 and 2.~\jh{How do we derive this observation from the image?} In this setting, early stopping outperforms single-pass sampling~\jh{Which line is single pass sampling?}, demonstrating that confidence scores serve as an effective method for distributing computational resources under limited budgets.

\subsection{Can Other Confidence Querying Prompts Work Well?}
\label{sec:confideceprompt}
Since our confidence-based approach was trained using a specific confidence querying prompt, we explore whether alternative prompts can achieve similar performance during inference. This analysis is crucial for understanding the robustness of confidence querying prompts different from the training prompt.

\begin{table}[]
\small
\centering
\begin{tabularx}{\columnwidth}{X l c c}
\toprule
Dataset & Method          & Original & New \\ 
\midrule
        & Early Stopping  & 81.7            & 81.52$_{\pm 0.30}$ \\ 
MathQA  & ASC w/o conf.   & 81.9            & 81.80$_{\pm 0.21}$ \\ 
        & SC w/o conf.    & 82.1            & 81.63$_{\pm 0.20}$ \\ 
\midrule
        & Early Stopping  & 67.2            & 70.80$_{\pm 1.99}$ \\ 
Obj\_C. & ASC w/o conf.   & 74.8            & 74.07$_{\pm 1.03}$ \\ 
        & SC w/o conf.    & 74.4            & 73.40$_{\pm 0.75}$ \\ 
\midrule
        & Early Stopping  & 86.2            & 86.62$_{\pm 0.20}$ \\ 
ARC\_C. & ASC w/o conf.   & 86.6            & 86.63$_{\pm 0.05}$ \\ 
        & SC w/o conf.    & 86.4            & 86.35$_{\pm 0.25}$ \\ 
\bottomrule
\end{tabularx}
\caption{Accuracy comparison between the original prompt ``Is the answer correct? (Yes/No)'' and 6 alternative confidence querying prompts on three datasets of Llama-3.1B-Instruct-SC. Results are reported as mean$_{\pm \text{std}}$. We report the detailed results for each alternative prompt  respectively in Appendix~\ref{app:result_prompt}.}

% ~\jh{We report the detailed results for each alternative prompt ($I_2$ to $I_6$) respectively in Appendix xx.}}
\vspace{-15pt}
\label{tab:confidence_prompt_performance}
\end{table}

We evaluate 6 alternative prompts (listed in Appendix~\ref{app:query_prompt}) at inference time. 
Table~\ref{tab:confidence_prompt_performance} shows that despite training with a specific prompt, other prompts yield comparable performance across all datasets, with only minor variations. This suggests that our confidence querying approach is robust to prompt changes and our training framework improves model calibration rather than overfitting to a special prompt.
% making confidence scores more stable and reliable.

% \jh{Other possible analysis: (1) hyperparameter study? (2) varying the confidence querying prompt? (3) temperature used in sampling? }

\section{Related Work}

\subsection{Test-Time Scaling}
% \jh{I feel that there are a lot of work in test-time scaling in addition to the ones you mentioned. Can you check out this tutorial: https://cmu-l3.github.io/neurips2024-inference-tutorial/ and list a few highly-cited ones? Also, you can check out their slides on the website.}
\citet{snell2024scaling} studied optimal test-time compute allocation to significantly enhance efficiency.
Self-Enhanced tree search frameworks~\citep{bi2024forest, lample2022hypertreeproofsearchneural, koh2024treesearchlanguagemodel} aggregate multiple reasoning paths and employs sparse activation strategies. Beyond that, step-wise verifiers are leveraged to dynamically prune the search tree~\citep{wang2023selfconsistencyimproveschainthought, li2023makinglargelanguagemodels, lightman2023letsverifystepstep}.
Additionally, \citet{chen2024simple} developed a two-stage elimination-based approach where multiple candidates are iteratively refined through pairwise comparisons.
Combining different versions of the same query can also improve the final performance~\cite{Huang2024DivideRA}.
Scaling~\citep{chen2025sets, welleck2022generatingsequenceslearningselfcorrect, wang2024theoreticalunderstandingselfcorrectionincontext, chen2023teachinglargelanguagemodels, madaan2023selfrefineiterativerefinementselffeedback, aggarwal2024alphaverusbootstrappingformallyverified} that iteratively refines model outputs, leading to improved performance in complex tasks.
\citet{muennighoff2025s1} proposed \textit{s1}, a simple test-time scaling method that enforces a budget constraint on inference length to optimize computational resource utilization. 

\subsection{Model Calibration}

Model calibration aims to align a model’s confidence with its accuracy. LLMs often exhibit overconfidence~\citep{tian2023just, chen2024reconfidencing, xiong2023can, achiam2023gpt}. Prior research has explored scaling-based methods~\citep{deng2023great, guo2017calibration, zhang2020mix} and nonparametric techniques like binning~\citep{zadrozny2001obtaining}. More recent work has introduced verbalized confidence, prompting models to directly output confidence scores~\citep{lin2022teaching}. Most studies focus on pre-trained and instruction-tuned LLMs~\citep{lin2022teaching, han2024enhancing}, others investigate RLHF-trained LLMs and propose calibration through prompting strategies~\citep{xiong2023can, tian2023just}. Reinforcement learning has also been leveraged for calibration~\citep{xu2024sayself, tao2024trust}, aligning closely with our study. A more calibrated reward model can also help model calibration by PPO framework~\citep{Leng2024TamingOI}.

\subsection{LLM Verifier}
Recently, various LLM verifiers are developed to enhance the reasoning capabilities of LLMs. Our approach is closely related to LLM-based verifiers, as both aim to evaluate whether a generated response meets correctness criteria.
\citet{Lightman2023LetsVS} trained verifiers that assess the correctness of generated solutions, enhancing the selection of accurate responses.
LLM-as-a-Judge~\citep{Zheng2023JudgingLW} employs large language models to adjudicate between multiple generated outputs based on learned preferences.
\citet{Zhang2024GenerativeVR} trained verifiers using next-token prediction to enhance reasoning performance in large language models.
GenRM~\citep{Mahan2024GenerativeRM} is an iterative algorithm that trains large language models on self-generated reasoning traces to align synthetic preference labels with human judgments.

\section{Conclusion}

We improve the efficiency of test-time scaling methods in LLMs with reliable confidence estimation.
Our Self-Calibration enhances LLM confidence estimation in one forward pass, without requiring any labeled data. 
We then propose efficient test-time scaling by dynamically adjusting sampling strategies based on calibrated confidence scores, such as Early-Stopping for Best-of-N and Self-Consistency with calibrated confidence. Experiments show that our approaches consistently outperform baselines under the same sample budget. Our findings suggest that reliable confidence estimation and dynamic sampling can substantially enhance the effectiveness and efficiency of test-time scaling approaches.

\section*{Acknowledgment}
This research was supported in part by the NVIDIA Academic Grant Program.

% \section*{Limitation}
% Despite the promising results demonstrated in this study, our work has certain limitations. Due to computational resource constraints, we were unable to conduct experiments on larger models, which might provide further insights into the scalability and generalizability of our approach. Additionally, the limited resources restricted the extent of hyperparameter tuning, preventing a more exhaustive search for optimal configurations. Future work could address these limitations by leveraging more powerful hardware or distributed computing strategies to enable experiments on larger models and more comprehensive hyperparameter optimization.

\bibliography{custom}

\begin{thebibliography}{71}
\providecommand{\natexlab}[1]{#1}

\bibitem[{Achiam et~al.(2023)Achiam, Adler, Agarwal, Ahmad, Akkaya, Aleman, Almeida, Altenschmidt, Altman, Anadkat et~al.}]{achiam2023gpt}
Josh Achiam, Steven Adler, Sandhini Agarwal, Lama Ahmad, Ilge Akkaya, Florencia~Leoni Aleman, Diogo Almeida, Janko Altenschmidt, Sam Altman, Shyamal Anadkat, et~al. 2023.
\newblock Gpt-4 technical report.
\newblock \emph{ArXiv preprint}, abs/2303.08774.

\bibitem[{Aggarwal et~al.(2023)Aggarwal, Madaan, Yang, and Mausam}]{Aggarwal2023LetsSS}
Pranjal Aggarwal, Aman Madaan, Yiming Yang, and Mausam. 2023.
\newblock Let's sample step by step: Adaptive-consistency for efficient reasoning and coding with llms.
\newblock In \emph{Conference on Empirical Methods in Natural Language Processing}.

\bibitem[{Aggarwal et~al.(2024)Aggarwal, Parno, and Welleck}]{aggarwal2024alphaverusbootstrappingformallyverified}
Pranjal Aggarwal, Bryan Parno, and Sean Welleck. 2024.
\newblock Alphaverus: Bootstrapping formally verified code generation through self-improving translation and treefinement.

\bibitem[{Amini et~al.(2024)Amini, Vieira, and Cotterell}]{Amini2024VariationalBA}
Afra Amini, Tim Vieira, and Ryan Cotterell. 2024.
\newblock Variational best-of-n alignment.
\newblock \emph{ArXiv preprint}, abs/2407.06057.

\bibitem[{Amini et~al.(2019)Amini, Gabriel, Lin, Koncel-Kedziorski, Choi, and Hajishirzi}]{Amini2019MathQATI}
Aida Amini, Saadia Gabriel, Shanchuan Lin, Rik Koncel-Kedziorski, Yejin Choi, and Hannaneh Hajishirzi. 2019.
\newblock \href {https://doi.org/10.18653/v1/N19-1245} {{M}ath{QA}: Towards interpretable math word problem solving with operation-based formalisms}.
\newblock In \emph{Proc. of NAACL-HLT}, pages 2357--2367, Minneapolis, Minnesota. Association for Computational Linguistics.

\bibitem[{Bi et~al.(2024)}]{bi2024forest}
Bin Bi et~al. 2024.
\newblock Forest-of-thought: Scaling test-time compute for enhancing llm reasoning.
\newblock \emph{ArXiv preprint}, abs/2412.09078.

\bibitem[{Brown et~al.(2024)Brown, Juravsky, Ehrlich, Clark, Le, R'e, and Mirhoseini}]{Brown2024LargeLM}
Bradley Brown, Jordan Juravsky, Ryan Ehrlich, Ronald Clark, Quoc~V. Le, Christopher R'e, and Azalia Mirhoseini. 2024.
\newblock Large language monkeys: Scaling inference compute with repeated sampling.
\newblock \emph{ArXiv preprint}, abs/2407.21787.

\bibitem[{Chen et~al.(2025{\natexlab{a}})Chen, Ren, Chen, Yang, Sun, and Arik}]{Chen2025SETSLS}
Jiefeng Chen, Jie Ren, Xinyun Chen, Chengrun Yang, Ruoxi Sun, and Sercan~{\"O}. Arik. 2025{\natexlab{a}}.
\newblock Sets: Leveraging self-verification and self-correction for improved test-time scaling.

\bibitem[{Chen et~al.(2025{\natexlab{b}})Chen, Ren, Chen, Yang, Sun, and Arık}]{chen2025sets}
Jiefeng Chen, Jie Ren, Xinyun Chen, Chengrun Yang, Ruoxi Sun, and Sercan~Ö Arık. 2025{\natexlab{b}}.
\newblock Sets: Leveraging self-verification and self-correction for improved test-time scaling.
\newblock \emph{ArXiv preprint}, abs/2501.19306.

\bibitem[{Chen et~al.(2024{\natexlab{a}})Chen, Perez-Lebel, Suchanek, and Varoquaux}]{chen2024reconfidencing}
Lihu Chen, Alexandre Perez-Lebel, Fabian~M Suchanek, and Ga{\"e}l Varoquaux. 2024{\natexlab{a}}.
\newblock Reconfidencing llms from the grouping loss perspective.
\newblock \emph{ArXiv preprint}, abs/2402.04957.

\bibitem[{Chen et~al.(2024{\natexlab{b}})Chen, Xu, Liang, He, Pang, Yu, Song, Liu, Zhou, Zhang, Wang, Tu, Mi, and Yu}]{Chen2024DoNT}
Xingyu Chen, Jiahao Xu, Tian Liang, Zhiwei He, Jianhui Pang, Dian Yu, Linfeng Song, Qiuzhi Liu, Mengfei Zhou, Zhuosheng Zhang, Rui Wang, Zhaopeng Tu, Haitao Mi, and Dong Yu. 2024{\natexlab{b}}.
\newblock Do not think that much for 2+3=? on the overthinking of o1-like llms.
\newblock \emph{ArXiv preprint}, abs/2412.21187.

\bibitem[{Chen et~al.(2023)Chen, Lin, Schärli, and Zhou}]{chen2023teachinglargelanguagemodels}
Xinyun Chen, Maxwell Lin, Nathanael Schärli, and Denny Zhou. 2023.
\newblock Teaching large language models to self-debug.

\bibitem[{Chen et~al.(2024{\natexlab{c}})Chen, Pan, Li, Ding, and Zhou}]{chen2024simple}
Yanxi Chen, Xuchen Pan, Yaliang Li, Bolin Ding, and Jingren Zhou. 2024{\natexlab{c}}.
\newblock A simple and provable scaling law for the test-time compute of large language models.
\newblock \emph{ArXiv preprint}, abs/2411.19477.

\bibitem[{Christiano et~al.(2017)Christiano, Leike, Brown, Martic, Legg, and Amodei}]{Christiano2017DeepRL}
Paul~F. Christiano, Jan Leike, Tom~B. Brown, Miljan Martic, Shane Legg, and Dario Amodei. 2017.
\newblock Deep reinforcement learning from human preferences.
\newblock In \emph{Advances in Neural Information Processing Systems 30: Annual Conference on Neural Information Processing Systems 2017, December 4-9, 2017, Long Beach, CA, {USA}}, pages 4299--4307.

\bibitem[{Clark et~al.(2018)Clark, Cowhey, Etzioni, Khot, Sabharwal, Schoenick, and Tafjord}]{allenai:arc}
Peter Clark, Isaac Cowhey, Oren Etzioni, Tushar Khot, Ashish Sabharwal, Carissa Schoenick, and Oyvind Tafjord. 2018.
\newblock Think you have solved question answering? try arc, the ai2 reasoning challenge.
\newblock \emph{ArXiv preprint}, abs/1803.05457.

\bibitem[{Cobbe et~al.(2021{\natexlab{a}})Cobbe, Kosaraju, Bavarian, Chen, Jun, Kaiser, Plappert, Tworek, Hilton, Nakano, Hesse, and Schulman}]{Cobbe2021TrainingVT}
Karl Cobbe, Vineet Kosaraju, Mohammad Bavarian, Mark Chen, Heewoo Jun, Lukasz Kaiser, Matthias Plappert, Jerry Tworek, Jacob Hilton, Reiichiro Nakano, Christopher Hesse, and John Schulman. 2021{\natexlab{a}}.
\newblock Training verifiers to solve math word problems.
\newblock \emph{ArXiv preprint}, abs/2110.14168.

\bibitem[{Cobbe et~al.(2021{\natexlab{b}})Cobbe, Kosaraju, Bavarian, Chen, Jun, Kaiser, Plappert, Tworek, Hilton, Nakano, Hesse, and Schulman}]{cobbe2021gsm8k}
Karl Cobbe, Vineet Kosaraju, Mohammad Bavarian, Mark Chen, Heewoo Jun, Lukasz Kaiser, Matthias Plappert, Jerry Tworek, Jacob Hilton, Reiichiro Nakano, Christopher Hesse, and John Schulman. 2021{\natexlab{b}}.
\newblock Training verifiers to solve math word problems.
\newblock \emph{ArXiv preprint}, abs/2110.14168.

\bibitem[{DeepSeek-AI(2025)}]{deepseekai2025deepseekr1incentivizingreasoningcapability}
DeepSeek-AI. 2025.
\newblock Deepseek-r1: Incentivizing reasoning capability in llms via reinforcement learning.

\bibitem[{Deng et~al.(2023)Deng, Xiong, and Hooi}]{deng2023great}
Ailin Deng, Miao Xiong, and Bryan Hooi. 2023.
\newblock Great models think alike: Improving model reliability via inter-model latent agreement.
\newblock \emph{ArXiv preprint}, abs/2305.01481.

\bibitem[{Dong et~al.(2024)Dong, Xiong, Pang, Wang, Zhao, Zhou, Jiang, Sahoo, Xiong, and Zhang}]{dong2024rlhf}
Hanze Dong, Wei Xiong, Bo~Pang, Haoxiang Wang, Han Zhao, Yingbo Zhou, Nan Jiang, Doyen Sahoo, Caiming Xiong, and Tong Zhang. 2024.
\newblock Rlhf workflow: From reward modeling to online rlhf.
\newblock \emph{ArXiv preprint}, abs/2405.07863.

\bibitem[{Dubey et~al.(2024)Dubey, Jauhri, Pandey, Kadian, Al-Dahle, Letman, Mathur, Schelten, and etc.}]{Dubey2024TheL3}
Abhimanyu Dubey, Abhinav Jauhri, Abhinav Pandey, Abhishek Kadian, Ahmad Al-Dahle, Aiesha Letman, Akhil Mathur, Alan Schelten, and etc. 2024.
\newblock The llama 3 herd of models.
\newblock \emph{ArXiv preprint}, abs/2407.21783.

\bibitem[{Guo et~al.(2017{\natexlab{a}})Guo, Pleiss, Sun, and Weinberger}]{Guo2017OnCO}
Chuan Guo, Geoff Pleiss, Yu~Sun, and Kilian~Q. Weinberger. 2017{\natexlab{a}}.
\newblock On calibration of modern neural networks.
\newblock In \emph{Proc. of ICML}, volume~70 of \emph{Proceedings of Machine Learning Research}, pages 1321--1330. {PMLR}.

\bibitem[{Guo et~al.(2017{\natexlab{b}})Guo, Pleiss, Sun, and Weinberger}]{guo2017calibration}
Chuan Guo, Geoff Pleiss, Yu~Sun, and Kilian~Q. Weinberger. 2017{\natexlab{b}}.
\newblock On calibration of modern neural networks.
\newblock In \emph{Proc. of ICML}, volume~70 of \emph{Proceedings of Machine Learning Research}, pages 1321--1330. {PMLR}.

\bibitem[{Han et~al.(2024)Han, Li, Chen, Shi, Du, Xiao, Liang, and Lin}]{han2024enhancing}
Haixia Han, Tingyun Li, Shisong Chen, Jie Shi, Chengyu Du, Yanghua Xiao, Jiaqing Liang, and Xin Lin. 2024.
\newblock Enhancing confidence expression in large language models through learning from past experience.
\newblock \emph{ArXiv preprint}, abs/2404.10315.

\bibitem[{Hendrycks and Gimpel(2017)}]{hendrycks2016baseline}
Dan Hendrycks and Kevin Gimpel. 2017.
\newblock A baseline for detecting misclassified and out-of-distribution examples in neural networks.
\newblock In \emph{Proc. of ICLR}. OpenReview.net.

\bibitem[{Huang et~al.(2024)Huang, Huang, and Huang}]{Huang2024DivideRA}
Chengsong Huang, Langlin Huang, and Jiaxin Huang. 2024.
\newblock Divide, reweight, and conquer: A logit arithmetic approach for in-context learning.
\newblock \emph{ArXiv preprint}, abs/2410.10074.

\bibitem[{Huang et~al.(2022)Huang, Gu, Hou, Wu, Wang, Yu, and Han}]{Huang2022LargeLM}
Jiaxin Huang, Shixiang~Shane Gu, Le~Hou, Yuexin Wu, Xuezhi Wang, Hongkun Yu, and Jiawei Han. 2022.
\newblock Large language models can self-improve.
\newblock \emph{ArXiv preprint}, abs/2210.11610.

\bibitem[{Kadavath et~al.(2022)Kadavath, Conerly, Askell, Henighan, Drain, Perez, Schiefer, Dodds, Dassarma, Tran-Johnson, Johnston, El-Showk, Jones, Elhage, Hume, Chen, Bai, Bowman, Fort, Ganguli, Hernandez, Jacobson, Kernion, Kravec, Lovitt, Ndousse, Olsson, Ringer, Amodei, Brown, Clark, Joseph, Mann, McCandlish, Olah, and Kaplan}]{Kadavath2022LanguageM}
Saurav Kadavath, Tom Conerly, Amanda Askell, Tom Henighan, Dawn Drain, Ethan Perez, Nicholas Schiefer, Zachary Dodds, Nova Dassarma, Eli Tran-Johnson, Scott Johnston, Sheer El-Showk, Andy Jones, Nelson Elhage, Tristan Hume, Anna Chen, Yuntao Bai, Sam Bowman, Stanislav Fort, Deep Ganguli, Danny Hernandez, Josh Jacobson, John Kernion, Shauna Kravec, Liane Lovitt, Kamal Ndousse, Catherine Olsson, Sam Ringer, Dario Amodei, Tom~B. Brown, Jack Clark, Nicholas Joseph, Benjamin Mann, Sam McCandlish, Christopher Olah, and Jared Kaplan. 2022.
\newblock Language models (mostly) know what they know.
\newblock \emph{ArXiv preprint}, abs/2207.05221.

\bibitem[{Koh et~al.(2024)Koh, McAleer, Fried, and Salakhutdinov}]{koh2024treesearchlanguagemodel}
Jing~Yu Koh, Stephen McAleer, Daniel Fried, and Ruslan Salakhutdinov. 2024.
\newblock Tree search for language model agents.

\bibitem[{Lample et~al.(2022)Lample, Lachaux, Lavril, Martinet, Hayat, Ebner, Rodriguez, and Lacroix}]{lample2022hypertreeproofsearchneural}
Guillaume Lample, Marie-Anne Lachaux, Thibaut Lavril, Xavier Martinet, Amaury Hayat, Gabriel Ebner, Aurélien Rodriguez, and Timothée Lacroix. 2022.
\newblock Hypertree proof search for neural theorem proving.

\bibitem[{Leng et~al.(2024)Leng, Huang, Zhu, and Huang}]{Leng2024TamingOI}
Jixuan Leng, Chengsong Huang, Banghua Zhu, and Jiaxin Huang. 2024.
\newblock Taming overconfidence in llms: Reward calibration in rlhf.
\newblock \emph{ArXiv preprint}, abs/2410.09724.

\bibitem[{Li et~al.(2022)Li, Lin, Zhang, Fu, Chen, Lou, and Chen}]{li2023makinglargelanguagemodels}
Yifei Li, Zeqi Lin, Shizhuo Zhang, Qiang Fu, Bei Chen, Jian-Guang Lou, and Weizhu Chen. 2022.
\newblock Making large language models better reasoners with step-aware verifier.

\bibitem[{Li et~al.(2024)Li, Yuan, Feng, Pan, Wang, Sun, Wang, and Li}]{Li2024EscapeSC}
Yiwei Li, Peiwen Yuan, Shaoxiong Feng, Boyuan Pan, Xinglin Wang, Bin Sun, Heda Wang, and Kan Li. 2024.
\newblock Escape sky-high cost: Early-stopping self-consistency for multi-step reasoning.
\newblock \emph{ArXiv preprint}, abs/2401.10480.

\bibitem[{Lightman et~al.(2023{\natexlab{a}})Lightman, Kosaraju, Burda, Edwards, Baker, Lee, Leike, Schulman, Sutskever, and Cobbe}]{lightman2023letsverifystepstep}
Hunter Lightman, Vineet Kosaraju, Yura Burda, Harri Edwards, Bowen Baker, Teddy Lee, Jan Leike, John Schulman, Ilya Sutskever, and Karl Cobbe. 2023{\natexlab{a}}.
\newblock Let's verify step by step.

\bibitem[{Lightman et~al.(2023{\natexlab{b}})Lightman, Kosaraju, Burda, Edwards, Baker, Lee, Leike, Schulman, Sutskever, and Cobbe}]{Lightman2023LetsVS}
Hunter Lightman, Vineet Kosaraju, Yura Burda, Harrison Edwards, Bowen Baker, Teddy Lee, Jan Leike, John Schulman, Ilya Sutskever, and Karl Cobbe. 2023{\natexlab{b}}.
\newblock Let's verify step by step.
\newblock \emph{ArXiv preprint}, abs/2305.20050.

\bibitem[{Lin et~al.(2022)Lin, Hilton, and Evans}]{lin2022teaching}
Stephanie Lin, Jacob Hilton, and Owain Evans. 2022.
\newblock Teaching models to express their uncertainty in words.
\newblock \emph{ArXiv preprint}, abs/2205.14334.

\bibitem[{Liu et~al.(2020)Liu, Cui, Liu, Huang, Wang, and Zhang}]{Liu2020LogiQAAC}
Jian Liu, Leyang Cui, Hanmeng Liu, Dandan Huang, Yile Wang, and Yue Zhang. 2020.
\newblock \href {https://doi.org/10.24963/ijcai.2020/501} {Logiqa: {A} challenge dataset for machine reading comprehension with logical reasoning}.
\newblock In \emph{Proceedings of the Twenty-Ninth International Joint Conference on Artificial Intelligence, {IJCAI} 2020}, pages 3622--3628. ijcai.org.

\bibitem[{Madaan et~al.(2023)Madaan, Tandon, Gupta, Hallinan, Gao, Wiegreffe, Alon, Dziri, Prabhumoye, Yang, Gupta, Majumder, Hermann, Welleck, Yazdanbakhsh, and Clark}]{madaan2023selfrefineiterativerefinementselffeedback}
Aman Madaan, Niket Tandon, Prakhar Gupta, Skyler Hallinan, Luyu Gao, Sarah Wiegreffe, Uri Alon, Nouha Dziri, Shrimai Prabhumoye, Yiming Yang, Shashank Gupta, Bodhisattwa~Prasad Majumder, Katherine Hermann, Sean Welleck, Amir Yazdanbakhsh, and Peter Clark. 2023.
\newblock Self-refine: Iterative refinement with self-feedback.

\bibitem[{Mahan et~al.(2024)Mahan, Phung, Rafailov, Blagden, nathan lile, Castricato, Franken, Finn, and Albalak}]{Mahan2024GenerativeRM}
Dakota Mahan, Duy Phung, Rafael Rafailov, Chase Blagden, nathan lile, Louis Castricato, Jan-Philipp Franken, Chelsea Finn, and Alon Albalak. 2024.
\newblock Generative reward models.
\newblock \emph{ArXiv preprint}, abs/2410.12832.

\bibitem[{Mihaylov et~al.(2018)Mihaylov, Clark, Khot, and Sabharwal}]{Mihaylov2018CanAS}
Todor Mihaylov, Peter Clark, Tushar Khot, and Ashish Sabharwal. 2018.
\newblock \href {https://doi.org/10.18653/v1/D18-1260} {Can a suit of armor conduct electricity? a new dataset for open book question answering}.
\newblock In \emph{Proc. of EMNLP}, pages 2381--2391, Brussels, Belgium. Association for Computational Linguistics.

\bibitem[{Muennighoff et~al.(2025)Muennighoff, Yang, Shi, Li, Fei-Fei, Hajishirzi, Zettlemoyer, Liang, Candès, and Hashimoto}]{muennighoff2025s1}
Niklas Muennighoff, Zitong Yang, Weijia Shi, Xiang~Lisa Li, Li~Fei-Fei, Hannaneh Hajishirzi, Luke Zettlemoyer, Percy Liang, Emmanuel Candès, and Tatsunori Hashimoto. 2025.
\newblock s1: Simple test-time scaling.
\newblock \emph{ArXiv preprint}, abs/2501.19393.

\bibitem[{Patel et~al.(2021)Patel, Bhattamishra, and Goyal}]{patel-etal-2021-nlp}
Arkil Patel, Satwik Bhattamishra, and Navin Goyal. 2021.
\newblock \href {https://doi.org/10.18653/v1/2021.naacl-main.168} {Are {NLP} models really able to solve simple math word problems?}
\newblock In \emph{Proceedings of the 2021 Conference of the North American Chapter of the Association for Computational Linguistics: Human Language Technologies}, pages 2080--2094, Online. Association for Computational Linguistics.

\bibitem[{Pope et~al.(2022)Pope, Douglas, Chowdhery, Devlin, Bradbury, Levskaya, Heek, Xiao, Agrawal, and Dean}]{Pope2022EfficientlyST}
Reiner Pope, Sholto Douglas, Aakanksha Chowdhery, Jacob Devlin, James Bradbury, Anselm Levskaya, Jonathan Heek, Kefan Xiao, Shivani Agrawal, and Jeff Dean. 2022.
\newblock Efficiently scaling transformer inference.
\newblock \emph{ArXiv preprint}, abs/2211.05102.

\bibitem[{Sakaguchi et~al.(2019)Sakaguchi, Bras, Bhagavatula, and Choi}]{Sakaguchi2019WinoGrande}
Keisuke Sakaguchi, Ronan~Le Bras, Chandra Bhagavatula, and Yejin Choi. 2019.
\newblock Winogrande.
\newblock \emph{Communications of the ACM}, 64:99 -- 106.

\bibitem[{Snell et~al.(2024{\natexlab{a}})Snell, Lee, Xu, and Kumar}]{Snell2024ScalingLT}
Charlie Snell, Jaehoon Lee, Kelvin Xu, and Aviral Kumar. 2024{\natexlab{a}}.
\newblock Scaling llm test-time compute optimally can be more effective than scaling model parameters.
\newblock \emph{ArXiv preprint}, abs/2408.03314.

\bibitem[{Snell et~al.(2024{\natexlab{b}})}]{snell2024scaling}
Charlie Snell et~al. 2024{\natexlab{b}}.
\newblock Scaling llm test-time compute optimally can be more effective than scaling model parameters.
\newblock \emph{ArXiv preprint}, abs/2408.03314.

\bibitem[{Suzgun et~al.(2022)Suzgun, Scales, Scharli, Gehrmann, Tay, Chung, Chowdhery, Le, Chi, Zhou, and Wei}]{Suzgun2022ChallengingBT}
Mirac Suzgun, Nathan Scales, Nathanael Scharli, Sebastian Gehrmann, Yi~Tay, Hyung~Won Chung, Aakanksha Chowdhery, Quoc~V. Le, Ed~H. Chi, Denny Zhou, and Jason Wei. 2022.
\newblock Challenging big-bench tasks and whether chain-of-thought can solve them.
\newblock In \emph{Annual Meeting of the Association for Computational Linguistics}.

\bibitem[{Talmor et~al.(2019)Talmor, Herzig, Lourie, and Berant}]{talmor-etal-2019-commonsenseqa}
Alon Talmor, Jonathan Herzig, Nicholas Lourie, and Jonathan Berant. 2019.
\newblock \href {https://doi.org/10.18653/v1/N19-1421} {{C}ommonsense{QA}: A question answering challenge targeting commonsense knowledge}.
\newblock In \emph{Proc. of NAACL-HLT}, pages 4149--4158, Minneapolis, Minnesota. Association for Computational Linguistics.

\bibitem[{Tao et~al.(2024)Tao, Yao, Ding, Xie, Cao, Sun, Gao, Shen, and Ding}]{tao2024trust}
Shuchang Tao, Liuyi Yao, Hanxing Ding, Yuexiang Xie, Qi~Cao, Fei Sun, Jinyang Gao, Huawei Shen, and Bolin Ding. 2024.
\newblock When to trust llms: Aligning confidence with response quality.
\newblock \emph{ArXiv preprint}, abs/2404.17287.

\bibitem[{Team(2024)}]{qwen2.5}
Qwen Team. 2024.
\newblock Qwen2.5: A party of foundation models.

\bibitem[{Tian et~al.(2023{\natexlab{a}})Tian, Mitchell, Zhou, Sharma, Rafailov, Yao, Finn, and Manning}]{Tian2023JustAF}
Katherine Tian, Eric Mitchell, Allan Zhou, Archit Sharma, Rafael Rafailov, Huaxiu Yao, Chelsea Finn, and Christopher~D. Manning. 2023{\natexlab{a}}.
\newblock Just ask for calibration: Strategies for eliciting calibrated confidence scores from language models fine-tuned with human feedback.
\newblock \emph{ArXiv preprint}, abs/2305.14975.

\bibitem[{Tian et~al.(2023{\natexlab{b}})Tian, Mitchell, Zhou, Sharma, Rafailov, Yao, Finn, and Manning}]{tian2023just}
Katherine Tian, Eric Mitchell, Allan Zhou, Archit Sharma, Rafael Rafailov, Huaxiu Yao, Chelsea Finn, and Christopher~D Manning. 2023{\natexlab{b}}.
\newblock Just ask for calibration: Strategies for eliciting calibrated confidence scores from language models fine-tuned with human feedback.
\newblock \emph{ArXiv preprint}, abs/2305.14975.

\bibitem[{Wan et~al.(2024)Wan, Wu, Chen, and Li}]{Wan2024ReasoningAS}
Guangya Wan, Yuqi Wu, Jie Chen, and Sheng Li. 2024.
\newblock Reasoning aware self-consistency: Leveraging reasoning paths for efficient llm sampling.

\bibitem[{Wang et~al.(2024{\natexlab{a}})Wang, Song, Tian, Peng, Jin, Mi, Su, and Yu}]{Wang2024SelfConsistencyBC}
Ante Wang, Linfeng Song, Ye~Tian, Baolin Peng, Lifeng Jin, Haitao Mi, Jinsong Su, and Dong Yu. 2024{\natexlab{a}}.
\newblock Self-consistency boosts calibration for math reasoning.
\newblock In \emph{Conference on Empirical Methods in Natural Language Processing}.

\bibitem[{Wang et~al.(2022{\natexlab{a}})Wang, Wei, Schuurmans, Le, Chi, Narang, Chowdhery, and Zhou}]{wang2023selfconsistencyimproveschainthought}
Xuezhi Wang, Jason Wei, Dale Schuurmans, Quoc Le, Ed~Chi, Sharan Narang, Aakanksha Chowdhery, and Denny Zhou. 2022{\natexlab{a}}.
\newblock Self-consistency improves chain of thought reasoning in language models.

\bibitem[{Wang et~al.(2022{\natexlab{b}})Wang, Wei, Schuurmans, Le, Chi, and Zhou}]{Wang2022SelfConsistencyIC}
Xuezhi Wang, Jason Wei, Dale Schuurmans, Quoc Le, Ed~H. Chi, and Denny Zhou. 2022{\natexlab{b}}.
\newblock Self-consistency improves chain of thought reasoning in language models.
\newblock \emph{ArXiv preprint}, abs/2203.11171.

\bibitem[{Wang et~al.(2024{\natexlab{b}})Wang, Wu, Wei, Jegelka, and Wang}]{wang2024theoreticalunderstandingselfcorrectionincontext}
Yifei Wang, Yuyang Wu, Zeming Wei, Stefanie Jegelka, and Yisen Wang. 2024{\natexlab{b}}.
\newblock A theoretical understanding of self-correction through in-context alignment.

\bibitem[{Wei et~al.(2022)Wei, Wang, Schuurmans, Bosma, Chi, Xia, Le, and Zhou}]{Wei2022ChainOT}
Jason Wei, Xuezhi Wang, Dale Schuurmans, Maarten Bosma, Ed~H. Chi, F.~Xia, Quoc Le, and Denny Zhou. 2022.
\newblock Chain of thought prompting elicits reasoning in large language models.
\newblock \emph{ArXiv preprint}, abs/2201.11903.

\bibitem[{Welbl et~al.(2017)Welbl, Liu, and Gardner}]{Welbl2017CrowdsourcingMC}
Johannes Welbl, Nelson~F. Liu, and Matt Gardner. 2017.
\newblock \href {https://doi.org/10.18653/v1/W17-4413} {Crowdsourcing multiple choice science questions}.
\newblock In \emph{Proceedings of the 3rd Workshop on Noisy User-generated Text}, pages 94--106, Copenhagen, Denmark. Association for Computational Linguistics.

\bibitem[{Welleck et~al.(2022)Welleck, Lu, West, Brahman, Shen, Khashabi, and Choi}]{welleck2022generatingsequenceslearningselfcorrect}
Sean Welleck, Ximing Lu, Peter West, Faeze Brahman, Tianxiao Shen, Daniel Khashabi, and Yejin Choi. 2022.
\newblock Generating sequences by learning to self-correct.

\bibitem[{Wu et~al.(2024)Wu, Sun, Li, Welleck, and Yang}]{Wu2024InferenceSL}
Yangzhen Wu, Zhiqing Sun, Shanda Li, Sean Welleck, and Yiming Yang. 2024.
\newblock Inference scaling laws: An empirical analysis of compute-optimal inference for problem-solving with language models.

\bibitem[{Xiong et~al.(2023)Xiong, Hu, Lu, Li, Fu, He, and Hooi}]{xiong2023can}
Miao Xiong, Zhiyuan Hu, Xinyang Lu, Yifei Li, Jie Fu, Junxian He, and Bryan Hooi. 2023.
\newblock Can llms express their uncertainty? an empirical evaluation of confidence elicitation in llms.
\newblock \emph{ArXiv preprint}, abs/2306.13063.

\bibitem[{Xu et~al.(2024)Xu, Wu, Diao, Liu, Wang, Chen, and Gao}]{xu2024sayself}
Tianyang Xu, Shujin Wu, Shizhe Diao, Xiaoze Liu, Xingyao Wang, Yangyi Chen, and Jing Gao. 2024.
\newblock Sayself: Teaching llms to express confidence with self-reflective rationales.
\newblock \emph{ArXiv preprint}, abs/2405.20974.

\bibitem[{Yao et~al.(2023)Yao, Yu, Zhao, Shafran, Griffiths, Cao, and Narasimhan}]{Yao2023TreeOT}
Shunyu Yao, Dian Yu, Jeffrey Zhao, Izhak Shafran, Thomas~L. Griffiths, Yuan Cao, and Karthik Narasimhan. 2023.
\newblock Tree of thoughts: Deliberate problem solving with large language models.
\newblock \emph{ArXiv preprint}, abs/2305.10601.

\bibitem[{Yu et~al.(2020)Yu, Jiang, Dong, and Feng}]{Yu2020ReClorAR}
Weihao Yu, Zihang Jiang, Yanfei Dong, and Jiashi Feng. 2020.
\newblock Reclor: {A} reading comprehension dataset requiring logical reasoning.
\newblock In \emph{Proc. of ICLR}. OpenReview.net.

\bibitem[{Zadrozny and Elkan(2001)}]{zadrozny2001obtaining}
Bianca Zadrozny and Charles Elkan. 2001.
\newblock Obtaining calibrated probability estimates from decision trees and naive bayesian classifiers.
\newblock In \emph{Proceedings of the Eighteenth International Conference on Machine Learning {(ICML} 2001), Williams College, Williamstown, MA, USA, June 28 - July 1, 2001}, pages 609--616. Morgan Kaufmann.

\bibitem[{Zhang et~al.(2020)Zhang, Kailkhura, and Han}]{zhang2020mix}
Jize Zhang, Bhavya Kailkhura, and Thomas~Yong{-}Jin Han. 2020.
\newblock Mix-n-match : Ensemble and compositional methods for uncertainty calibration in deep learning.
\newblock In \emph{Proc. of ICML}, volume 119 of \emph{Proceedings of Machine Learning Research}, pages 11117--11128. {PMLR}.

\bibitem[{Zhang et~al.(2024{\natexlab{a}})Zhang, Hosseini, Bansal, Kazemi, Kumar, and Agarwal}]{Zhang2024GenerativeVR}
Lunjun Zhang, Arian Hosseini, Hritik Bansal, Mehran Kazemi, Aviral Kumar, and Rishabh Agarwal. 2024{\natexlab{a}}.
\newblock Generative verifiers: Reward modeling as next-token prediction.
\newblock \emph{ArXiv preprint}, abs/2408.15240.

\bibitem[{Zhang et~al.(2024{\natexlab{b}})Zhang, Bao, and Huang}]{Zhang2024EDTIL}
Shimao Zhang, Yu~Bao, and Shujian Huang. 2024{\natexlab{b}}.
\newblock Edt: Improving large language models' generation by entropy-based dynamic temperature sampling.
\newblock \emph{ArXiv preprint}, abs/2403.14541.

\bibitem[{Zhang et~al.(2025)Zhang, Zheng, Wu, Zhang, Lin, Yu, Liu, Zhou, and Lin}]{prmlessons}
Zhenru Zhang, Chujie Zheng, Yangzhen Wu, Beichen Zhang, Runji Lin, Bowen Yu, Dayiheng Liu, Jingren Zhou, and Junyang Lin. 2025.
\newblock The lessons of developing process reward models in mathematical reasoning.
\newblock \emph{ArXiv preprint}, abs/2501.07301.

\bibitem[{Zheng et~al.(2023)Zheng, Chiang, Sheng, Zhuang, Wu, Zhuang, Lin, Li, Li, Xing, Zhang, Gonzalez, and Stoica}]{Zheng2023JudgingLW}
Lianmin Zheng, Wei-Lin Chiang, Ying Sheng, Siyuan Zhuang, Zhanghao Wu, Yonghao Zhuang, Zi~Lin, Zhuohan Li, Dacheng Li, Eric~P. Xing, Haotong Zhang, Joseph~E. Gonzalez, and Ion Stoica. 2023.
\newblock Judging llm-as-a-judge with mt-bench and chatbot arena.
\newblock \emph{ArXiv preprint}, abs/2306.05685.

\end{thebibliography}

\newpage
% \onecolumn
\appendix

\section{Prompts}
\label{app:prompt}
\subsection{System Prompt}
Here we show the system prompt to let the model generate responses for Chain-of-Thoughts and format for extracting the final results.

\begin{quote}
\ttfamily
For the following question, provide a step-by-step explanation of your thought process. \\
Use the format demonstrated below for your response. \\

\verb|`|\verb|`|\verb|`|Example Format: \\
\texttt{Explanation: <Your detailed explanation here, outlining how you arrived at your answer.>} \\
\texttt{Answer: <Insert your concise answer here, which should include a \{\textit{answer\_type}\} (e.g., \{\textit{demo}\})>} \\

Ensure that your response strictly adheres to this format. Explicitly include the words \texttt{'Explanation:'}, \texttt{'Answer:'}. \\
\end{quote}

The answer type includes ``option letter'' and ``number''.

\subsection{Dataset Prompts}
We show the prompts for each dataset in Table~\ref{tab:dataset-prompts}.
All datasets and models are open-sourced.
\begin{table*}[t]
\centering
\small
\begin{tabular}{@{}lp{10cm}@{}}
\toprule
\textbf{Dataset} & \textbf{Query Template} \\
\midrule

\textbf{gsm8k} &
\texttt{Question: \{question\}\textbackslash n}
\\[6pt]

\textbf{sciq} &
\texttt{Question: \{question\}\textbackslash nOptions:\textbackslash n\{options\_text\}\textbackslash n}
\\[6pt]

\textbf{commonsense\_qa} &
\texttt{Question: \{question\}\textbackslash nOptions:\textbackslash n\{options\_text\}\textbackslash n}
\\[6pt]

\textbf{winogrande} &
\texttt{Question: \{sentence\}\textbackslash nOptions:\textbackslash nA. \{option1\}\textbackslash nB. \{option2\}\textbackslash n}
\\[6pt]

\textbf{openbookqa} &
\texttt{Question: \{question\}\textbackslash nOptions:\textbackslash n\{options\_text\}\textbackslash n}
\\[6pt]

\textbf{reclor} &
\texttt{Passage:\textbackslash n\{passage\}\textbackslash n\textbackslash nQuestion: \{question\}\textbackslash n\textbackslash nOptions:\textbackslash n\{options\_text\}\textbackslash n}
\\[6pt]

\textbf{math\_qa} &
\texttt{Problem: \{problem\_text\}\textbackslash nOptions:\textbackslash n\{options\_block\}\textbackslash n}
\\[6pt]

\textbf{arc\_challenge} &
\texttt{Question: \{question\}\textbackslash nOptions:\textbackslash n\{options\_str\}\textbackslash n}
\\[6pt]

\textbf{arc\_easy} &
\texttt{Question: \{question\}\textbackslash nOptions:\textbackslash n\{options\_str\}\textbackslash n}
\\[6pt]

\textbf{logiqa} &
\texttt{Article:\textbackslash n\{context\}\textbackslash n\textbackslash nQuestion: \{question\}\textbackslash n\textbackslash nOptions:\textbackslash n\{options\_text\}\textbackslash n}
\\[6pt]

\textbf{svamp} &
\texttt{Question: \{Body + Question\}\textbackslash n}
\\[6pt]

\textbf{gpqa} &
\texttt{\{Question\}\textbackslash nOptions:\textbackslash n\{options\_text\}\textbackslash n}
\\[6pt]

\textbf{aqua\_rat} &
\texttt{Question: \{question\}\textbackslash nOptions:\textbackslash n\{options\_text\}\textbackslash n}
\\

\bottomrule
\end{tabular}
\caption{Query templates for each dataset .}

\label{tab:dataset-prompts}
\end{table*}

\section{Full Main Results}
\label{app:full_table}
Here we show the main results when sample budget = 4 in Table~\ref{tab:results1}.

\begin{table*}[h]
 \centering
 \small
 \setlength{\tabcolsep}{1.5pt}
 \resizebox{\textwidth}{!}{%
 \begin{tabular}{llll|lll|lll}
 \toprule
 & \multicolumn{3}{c}{Llama-3.1-8B-Instruct} & \multicolumn{3}{c}{Qwen2.5-7B-Instruct} & \multicolumn{3}{c}{DeepSeek-R1-Distill-1.5B} \\
 \cmidrule(lr){2-4} \cmidrule(lr){5-7} \cmidrule(lr){8-10}
 Methods & Obj\_C. & MathQA & ARC\_C. & Obj\_C. & MathQA & ARC\_C. & Obj\_C. & MathQA & ARC\_C. \\
 \midrule
 \rowcolor{gray!20} Pass@1 & 67.6 & 71.5 & 82.8 & 76.8 & 82.9 & 88.5 & 61.2 & 89.9 & 58.2 \\ 
 \midrule
 \multicolumn{10}{l}{\textit{Sample budget = 4}} \\
 SC & 72.0 & 78.8 & 85.7 & 78.8 & 85.7 & 90.0 & 63.6 & 91.4 & 63.0 \\
 SC w/ Conf.* & 72.4 (+0.4) & 81.8 (+3.0) & 86.4 (+0.7) & 78.2 (-0.6) & 86.5 (+0.8) & 89.8 (-0.2) & 60.8 (-3.2) & 90.6 (-0.8) & 62.6 (-0.4)\\
 \rowcolor{blue!10}SC w/ Conf. & 72.8 (+0.8) & \textbf{82.1} (+3.3) & 86.4 (+0.7) & 78.4 (-0.4) & 86.9 (+1.2) & 90.3 (+0.3) & \textbf{64.0} (+0.4) & 91.2 (-0.2) & 63.2 (+0.2) \\
 Best-of-N & 67.6 & 80.8 & 86.4 & 76.4 & 86.4 & 89.8 & 56.0 & 90.0 & 59.0 \\
 Early Stopping* & 65.6 (-2.0) & 81.2 (+0.4) & 86.1 (-0.3) & 76.0 (-0.4) & 86.6 (+0.2) & 89.6 (-0.2) & 55.2 (-0.8) & 90.5 (+0.5) & 58.8 (-0.2) \\
 \rowcolor{blue!10}Early Stopping & 67.2 (-0.4) & 81.7 (+0.9) & 86.2 (-0.2) & 78.4 (+2.0) & 87.1 (+0.7) & 90.1 (+0.3) & 56.0 (0.0) & 90.6 (+0.6) & 59.0 (0.0) \\
 ASC & 74.4 & 80.0 & 86.5 & 79.6 & 86.2 & \textbf{91.0} & 61.2 & 91.3 & 63.3 \\
 ASC w/ Conf.* & 73.2 (-1.2) & 81.7 (+1.7) & 86.5 (0.0) & 79.8 (+0.2) & 86.9 (+0.7) & 90.4 (-0.6) & 62.4 (+1.2) & 91.6 (+0.3) & 64.2 (+0.9) \\
 \rowcolor{blue!10}ASC w/ Conf. & \textbf{74.8} (+0.4) & 81.9 (+1.9) & \textbf{86.6} (+0.1) & \textbf{80.0} (+0.4) & \textbf{87.2} (+1.0) & 90.6 (-0.4) & {62.8} (+1.6) & \textbf{91.6} (+0.3) & \textbf{64.6} (+1.3) \\
 ESC & 72.0 & 78.6 & 85.8 & \textbf{80.0} & 86.9 & 89.6 & 58.0 & 91.2 & 63.0 \\
 RASC & 72.4 & 79.0 & 85.8 & \textbf{80.0} & 86.4 & 89.8 & 62.6 & 91.2 & 63.1 \\
 % \midrule
 % \midrule
 % \multicolumn{10}{l}{\textit{Sample budget = 16}} \\
 % SC & 76.0 & 81.0 & 87.1 & 81.2 & 86.3 & \textbf{91.2} & \textbf{70.8} & 91.6 & 65.6 \\
 % SC w/ Conf.* & \textbf{76.8} (+0.8) & 83.4 (+2.4)  & 87.4 (+0.3) & 80.8 (-0.4) & 87.5 (+1.2) & 90.5 (-0.7) & \textbf{70.8} (0.0) & \textbf{91.8} (+0.2) & 65.9 (+0.3) \\
 % \rowcolor{blue!10}SC w/ Conf.  & \textbf{76.8} (+0.8) & \textbf{83.6} (+2.6) & \textbf{87.7} (+0.6) & 81.2 (0.0) & \textbf{87.8} (+1.5) & 90.8 (-0.4) & \textbf{70.8} (0.0) & \textbf{91.8} (+0.2) & \textbf{66.5} (+0.9) \\
 % Best-of-N & 69.2 & 81.0 & 86.4 & 76.8 & 86.8 & 90.2 & 54.0 & 90.0 & 58.9 \\
 % Early Stopping* & \textbf{76.8} (+7.6) & 83.4 (+2.4) & 87.3 (+0.9) & 80.8 (+4.0) & 87.5 (+0.7) & 90.5 (+0.3) & 64.8 (+10.8) & 91.6 (+1.6) & 65.9 (+7.0)\\
 % \rowcolor{blue!10}Early Stopping & \textbf{76.8} (+7.6) & \textbf{83.6} (+2.6) & \textbf{87.7} (+1.3) & 81.2 (+4.4) & \textbf{87.8} (+1.0) & 90.8 (+0.6) & \textbf{70.8} (+16.8) & 91.6 (+1.6) & \textbf{66.5} (+7.6) \\
 % ASC & 74.8 & 80.0 & 86.5 & \textbf{81.6} & 86.2 & 90.6 & 70.4 & 91.6 & 64.3 \\
 % ASC w/ Conf.* & 74.8 (0.0) & 81.6 (+1.6) & 86.6 (+0.1) & \textbf{81.6} (0.0) & 86.9 (+0.7) & 90.4 (-0.2) & 70.4 (0.0) & 91.6 (0.0) & 64.7 (+0.4) \\
 % \rowcolor{blue!10}ASC w/ Conf. & 75.2 (+0.4) & 81.9 (+1.9) & 86.6 (+0.1) & \textbf{81.6} (0.0) & 87.2 (+1.0) & \textbf{91.2} (+0.6) & 70.4 (0.0) & \textbf{91.8} (+0.2) & 65.1 (+0.8) \\
 % ESC & 76.0 & 81.0 & 87.1 & 81.2 & 86.3 & {91.0} & \textbf{70.8} & 91.3 & 65.6 \\
 % RASC & 76.0 & 81.4 & 87.3 & 81.2 & 86.4 & 90.3 & \textbf{70.8} & 91.4 & 65.8 \\
 \bottomrule
 \end{tabular}
 }
 \caption{
 Accuracy comparison of different test-time scaling methods across three language models. The evaluation is conducted on three datasets: Obj\_C. (Object\_Counting), MathQA, and ARC\_C. (ARC\_Challenge). ``Sample budget'' refers to the average number of responses sampled per query. The improvements of confidence-augmented methods over their baselines are shown in parentheses. All methods use the same responses generated by Self-Calibration trained models, while methods marked with \textbf{*} use confidence scores from the vanilla model. 
 }
 % \vspace{-15pt}
 \label{tab:results1}
\end{table*}

When the sample budget is small, the model has limited opportunities to explore different reasoning paths. In this scenario, output variability is often high, and having an additional confidence signal (as in ASC w/ Conf.) is essential for filtering out noisy or incorrect responses. This confidence-augmented method helps select the most promising candidate under tight sampling constraints. 

However, when the sample budget increases, the model can generate more candidate solutions, which typically raises the chance of hitting the correct answer. In this setting, Early Stopping approach—especially when coupled with a high confidence threshold—can terminate as soon as it encounters a correct reasoning path.

\section{Full Results of Different Confidence Querying Prompts}
\label{app:confidence_query}

\subsection{Confidence Querying prompts}
\label{app:query_prompt}

We show the 6 confidence querying prompt we used in Sec.~\ref{sec:confideceprompt}.
\setlength\itemsep{0em}
\begin{itemize}[leftmargin=*]
    \item $I_1$: Is this the correct answer?
    \item $I_2$: Does this answer seem right?
    \item $I_3$: Is this the right answer?
    \item $I_4$: Is the given answer accurate?
    \item $I_5$: Would you say this answer is correct?
    \item $I_6$: Is this response correct?
\end{itemize}

\subsection{Results of Different Querying Prompts}\label{app:result_prompt}
In Table~\ref{tab:confidence_pr}, we show the results of different confidence querying prompts for tuned LLama-3.1-8B-Instruct.
\begin{table*}[h]
\small
\centering
\begin{tabular}{llccccccc}
\toprule
   Dataset              &     Method          & 1     & 2     & 3     & 4     & 5     & 6     & Original \\
\midrule
                 & Early Stopping & 81.7  & 81.4  & 81.7  & 81.3  & 81.1  & \textbf{81.9}  & 81.7  \\
\textbf{MathQA}  & ASC w/o conf.  & 81.9  & 81.9  & 81.8  & 81.8  & 81.4  & \textbf{82.0}  & 81.9  \\
                 & SC w/o conf.   & 81.5  & 81.4  & 81.5  & 81.7  & 81.9  & 81.8  & \textbf{82.1}  \\
\midrule
                 & Early Stopping & 70.0  & 71.6  & 69.6  & 68.0  & \textbf{73.6}  & 72.0  & 67.2  \\
\textbf{Object\_Counting} & ASC w/o conf.  & \textbf{73.6}  & 73.6  & 74.4  & 73.6  & \textbf{76.0}  & 73.2  & 74.8  \\
                 & SC w/o conf.   & 72.8  & \textbf{74.0}  & 73.2  & 72.4  & 74.4  & 73.6  & 72.8  \\
\midrule
                 & Early Stopping & \textbf{86.8}  & 86.4  & \textbf{86.8}  & 86.5  & \textbf{86.8}  & 86.4  & 86.2  \\
\textbf{ARC\_challenge}   & ASC w/o conf.  & \textbf{86.7}  & 86.6  & 86.6  & 86.6  & \textbf{86.7}  & 86.6  & 86.6  \\
                 & SC w/o conf.   & 86.3  & 86.1  & 86.1  & \textbf{86.7}  & 86.3  & 86.6  & 86.4  \\
\bottomrule
\end{tabular}
\caption{The results for different confidence querying prompt.}
\label{tab:confidence_pr}
\end{table*}

\section{Results for Different Sample Budgets}
\label{app:different_budgets}
Here, we show  the performance under different sample budgets of other datasets and models.

\begin{figure}
    \centering
\includegraphics[width=0.95\linewidth]{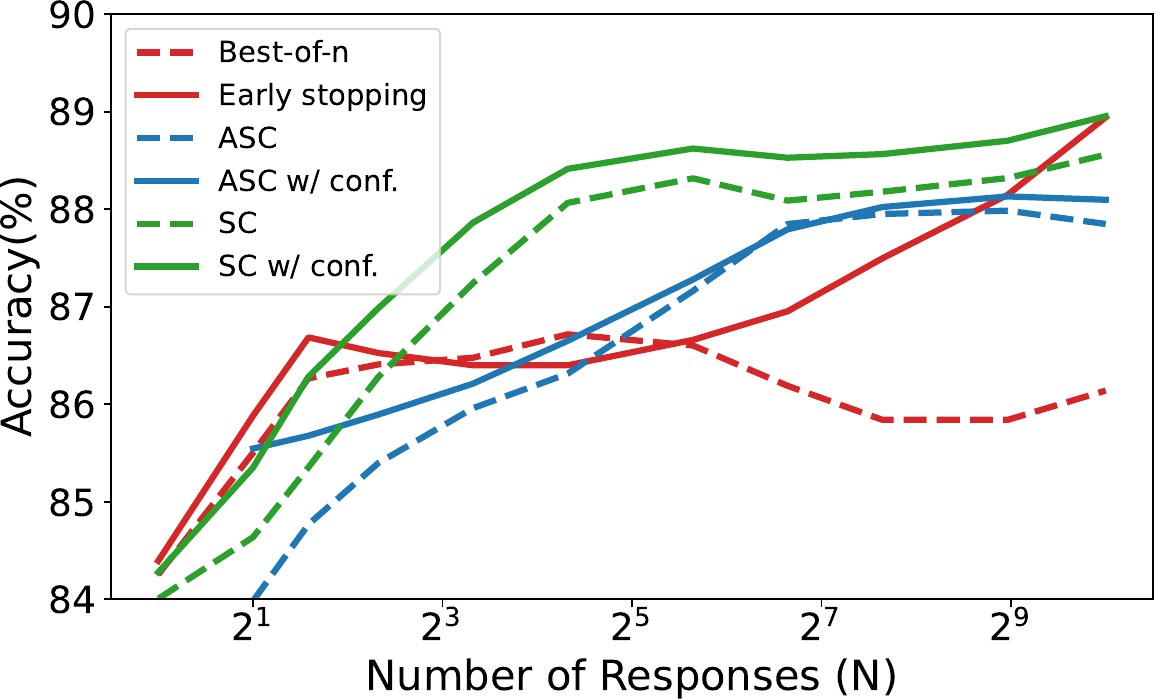}
    \caption{Performance comparison of different inference strategies on ARC\_Challenge using Self-Calibration
trained Llama-3.1-8B-Instruct. }
    \label{fig:enter-label}
\end{figure}

\begin{figure}
    \centering
\includegraphics[width=0.95\linewidth]{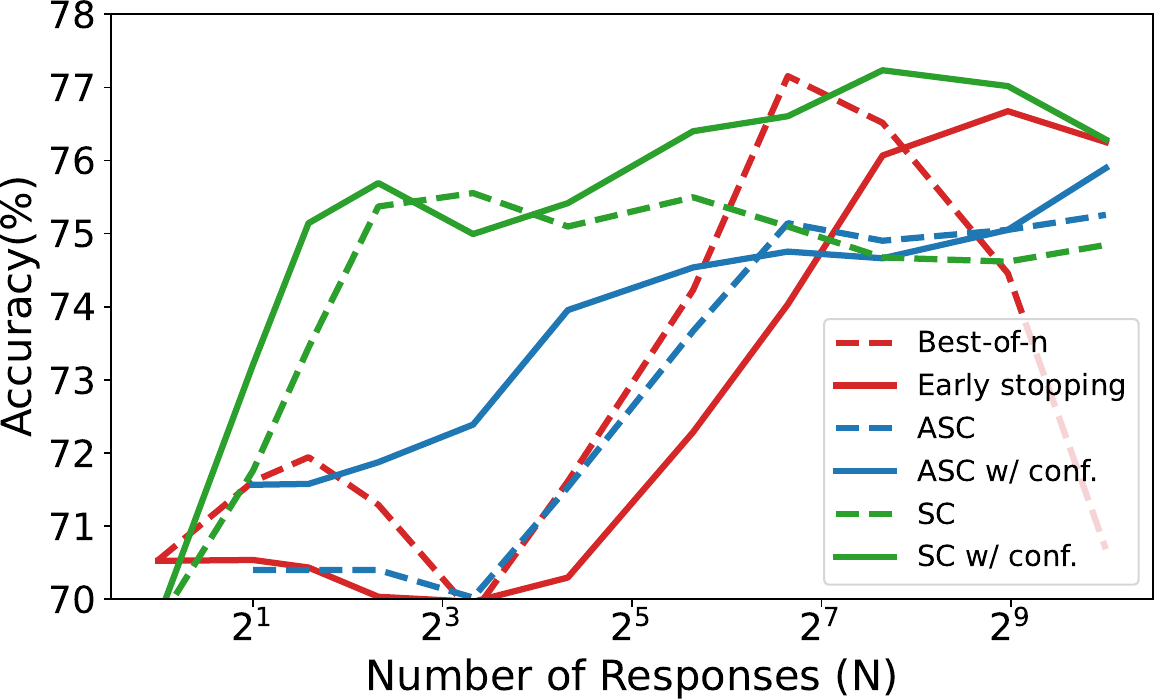}
    \caption{Performance comparison of different inference strategies on Object Counting using Self-Calibration
trained Llama-3.1-8B-Instruct. }
    \label{fig:enter-label}
\end{figure}

\begin{figure}
    \centering
\includegraphics[width=0.95\linewidth]{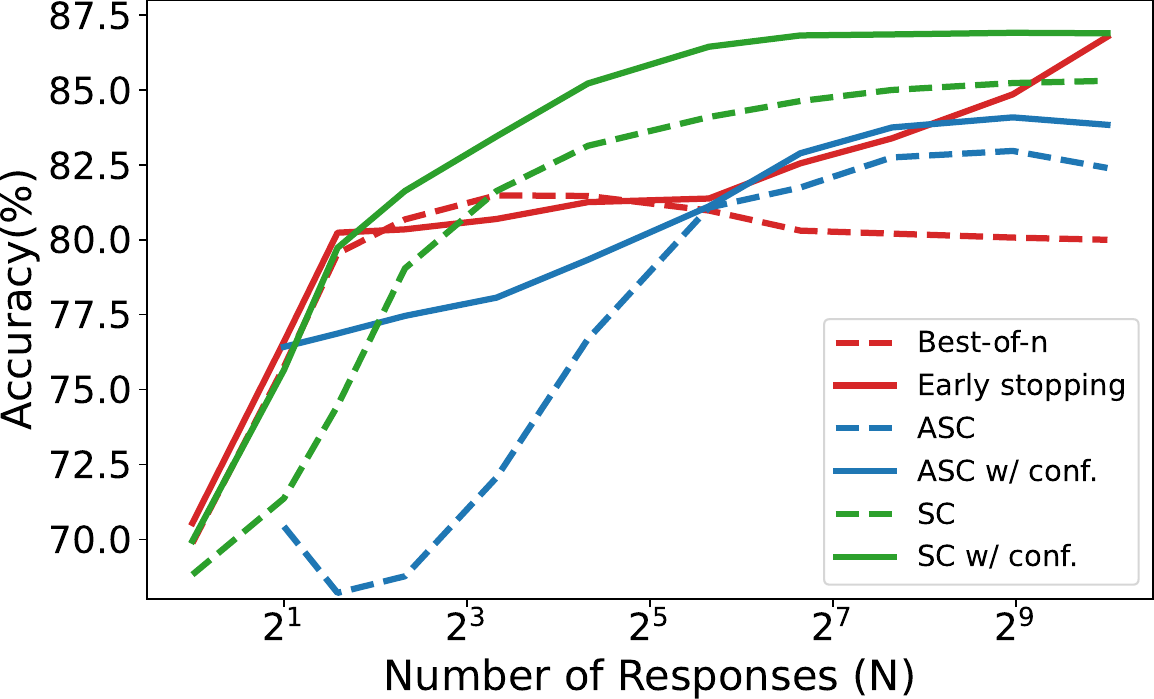}
    \caption{Performance comparison of different inference strategies on MathQA using Self-Calibration
trained Llama-3.1-8B-Instruct. }
    \label{fig:enter-label}
\end{figure}

\begin{figure}
    \centering
\includegraphics[width=0.95\linewidth]{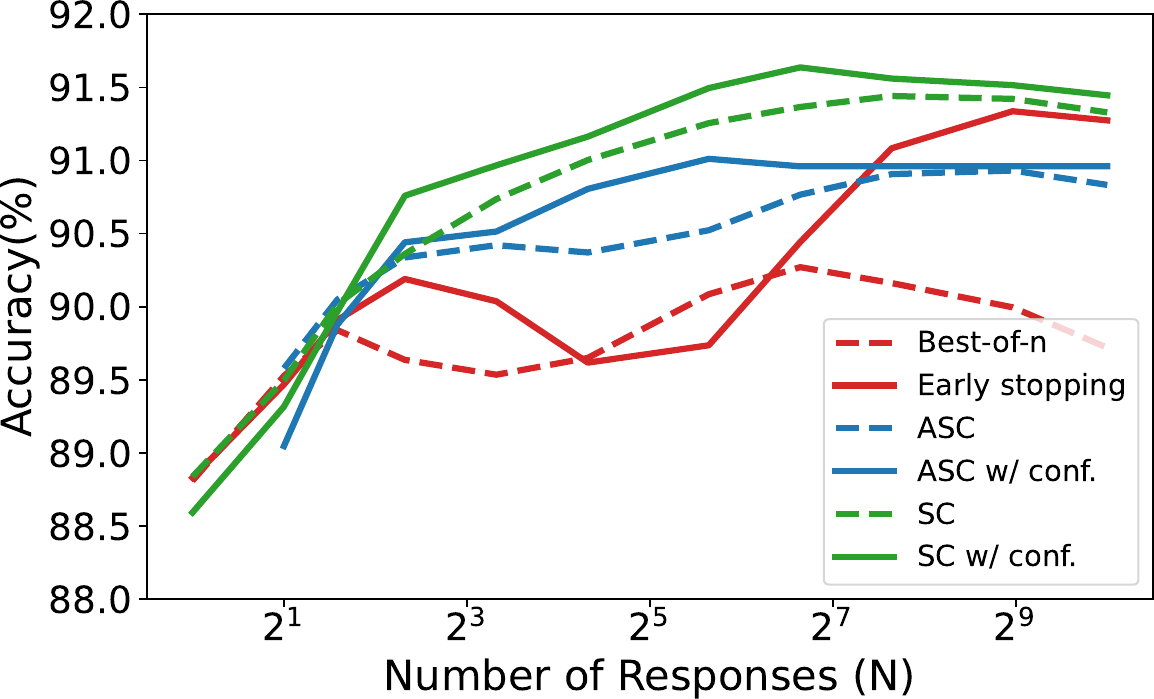}
    \caption{Performance comparison of different inference strategies on ARC\_Challenge using Self-Calibration
trained Qwen-2.5-7B-Instruction. }
    \label{fig:enter-label}
\end{figure}

\begin{figure}
    \centering
\includegraphics[width=0.95\linewidth]{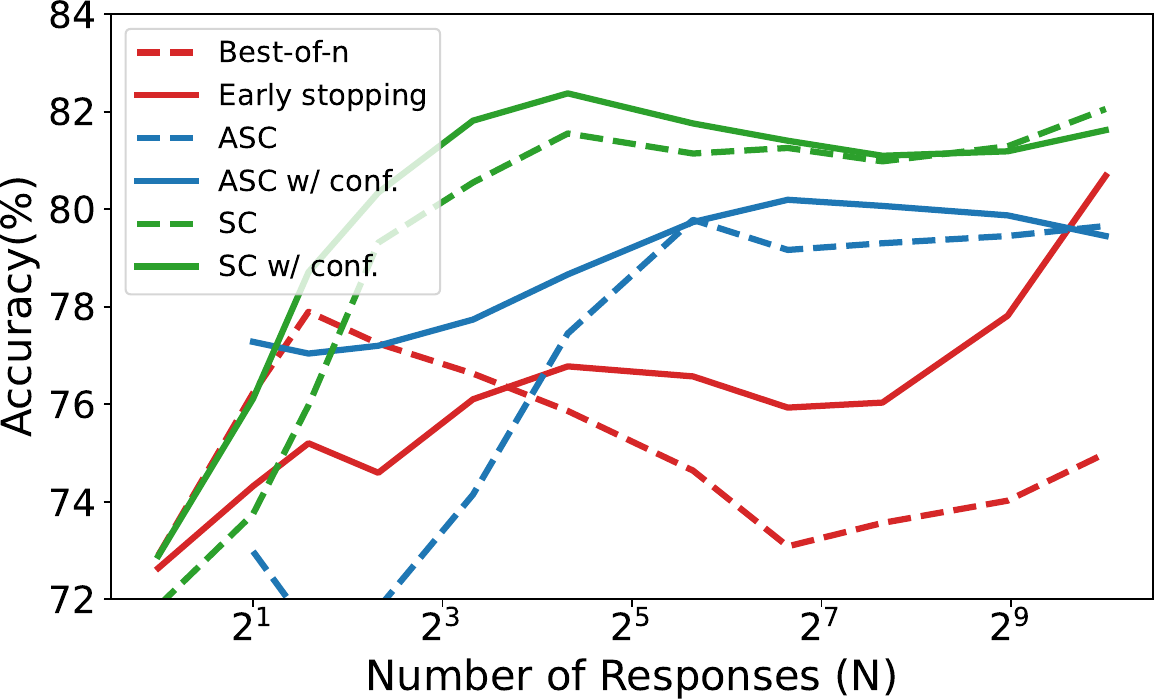}
    \caption{Performance comparison of different inference strategies on Object Counting using Self-Calibration
trained Qwen-2.5-7B-Instruction. }
    \label{fig:enter-label}
\end{figure}

\begin{figure}
    \centering
\includegraphics[width=0.95\linewidth]{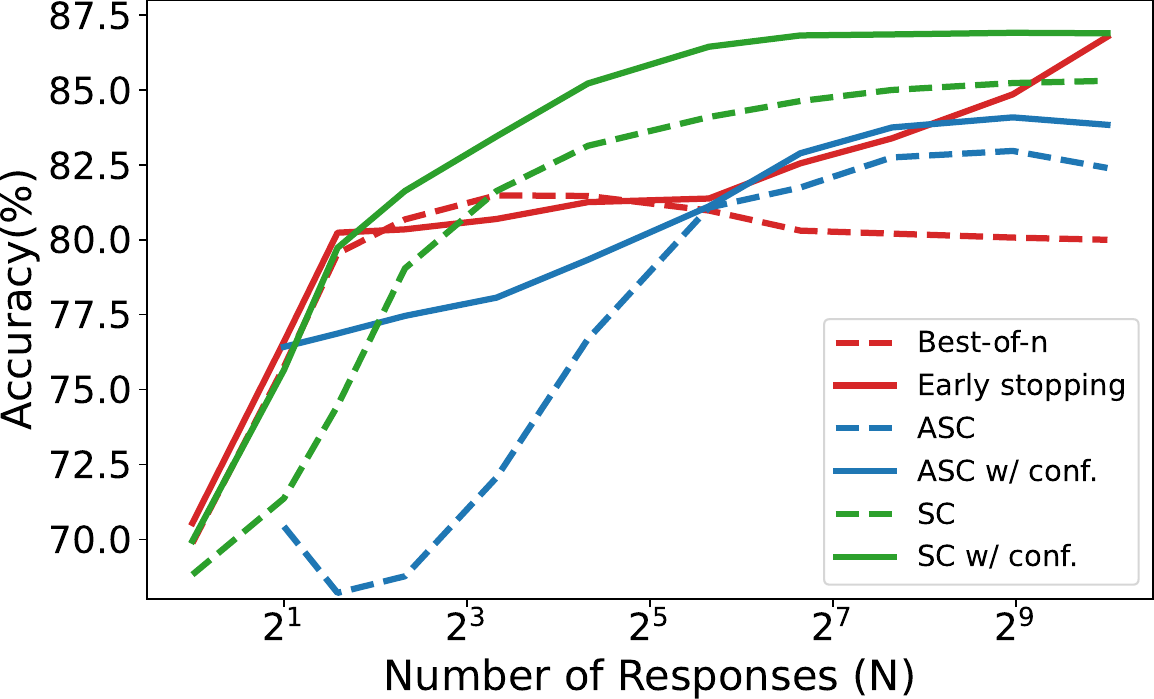}
    \caption{Performance comparison of different inference strategies on MathQA using Self-Calibration
trained Qwen-2.5-7B-Instruction. }
    \label{fig:enter-label}
\end{figure}

\section{Hyperparameters}
\label{app:hyperparameter}

This section details the hyperparameters used in our experiments. We categorize them into training data generation, training process, and response generation
\subsection{Training Data Generation}
When creating the datasets, we set the number of responses for each query $N=32$. For the parameter in dynamic temperature, we follow the default hyperparameter settings from the original paper: \(T_0=0.8\), \(M=0.8\), \(\gamma=1.0\), and \(\tau_0=0.001\).
\subsection{Training Process}
In the training objective, we set the threshold $\eta = 0.75$ to filter the response used in generation ability training and the weight $w=0.1$ to balance two losses.

In the training process, we use the AdamW optimizer with a learning rate of $5 \times 10^{-5}$. The total number of training samples is set to 100,000, while 1,000 samples are used for evaluation. We employ a batch size of 1 with gradient accumulation steps of 64 to simulate a larger effective batch size. The model is trained for 1 epoch.

For parameter-efficient fine-tuning, we apply LoRA with rank $r=32$, scaling factor $\alpha=16$, and dropout rate of $0.05$. In the whole training examples, the ratio of causal language modeling data is 0.7.
We train the model on multiple datasets with varying proportions of training and evaluation data. Specifically, GSM8K and SVAMP each contribute 15\% of the training and evaluation samples. SciQ, CommonsenseQA, Winogrande, OpenBookQA, ReClor, ARC-Easy, and LogiQA each contribute 5\% of the training and evaluation samples.

During the sample training data selection process, we ensure that the data is evenly distributed across different confidence intervals. This balancing strategy prevents overrepresentation of any specific confidence range, allowing the model to learn from a diverse set of samples. By maintaining an equal number of training examples in each confidence bin, we improve the robustness of confidence calibration and reduce potential biases in the learning process.

\subsection{Response Generation}
When generating the response, we set the temperature equals to 1.0.

\end{document}